\documentclass{elsart1p}
\usepackage{amsmath}
\usepackage{cases}
\usepackage{booktabs}
\usepackage{color}
\usepackage{footnote}
\usepackage{makecell}
\usepackage{float}
\usepackage{graphicx}
\usepackage{subfigure}
\usepackage{epsfig}
\usepackage{enumerate}
\usepackage{amssymb}
\usepackage{algorithm}
\usepackage{algpseudocode}
\usepackage{amsmath}

\newtheorem{theorem}{Theorem}

\begin{document}
\begin{frontmatter}
\title{Multi-task nonparallel support vector machine for classification}
\author{Zongmin Liu$^1$},
\author{Yitian Xu$^{2*}$}
\ead{xytshuxue@126.com, Tel.: +8610 62737077.}
\address{$^1$College of Information and Electrical Engineering, China Agricultural University, Beijing 100083, China}
\address{$^2$College of Science, China Agricultural University, Beijing 100083, China}
\begin{abstract}
Direct multi-task twin support vector machine (DMTSVM) explores the shared information between multiple correlated tasks, then it produces better generalization performance. However, it contains matrix inversion operation when solving the dual problems, so it costs much running time. Moreover, kernel trick cannot be directly utilized in the nonlinear case. To effectively avoid above problems, a novel multi-task nonparallel support vector machine (MTNPSVM) including linear and nonlinear cases is proposed in this paper. By introducing $\epsilon$-insensitive loss instead of square loss in DMTSVM, MTNPSVM effectively avoids matrix inversion operation and takes full advantage of the kernel trick. Theoretical implication of the model is further discussed. To further improve the computational efficiency, the alternating direction method of multipliers (ADMM) is employed when solving the dual problem. The computational complexity and convergence of the algorithm are provided. In addition, the property and sensitivity of the parameter in model are further explored. The experimental results on fifteen benchmark datasets and twelve image datasets demonstrate the validity of MTNPSVM in comparison with the state-of-the-art algorithms. Finally, it is applied to real Chinese Wine dataset, and also verifies its effectiveness.
\end{abstract}		
\end{frontmatter}
\section{Introduction}
In the single-task learning (STL) field, support vector machine (SVM) has attracted much academic attention in recent years due to its solid theoretical foundation and good performance, but it needs to deal with a large-scale problem, which leads to a low computational efficiency. Twin support vector machine (TWSVM) \cite{twsvm} proposed by Jayadeva et al. transforms a larger-scale problem in SVM into two small-scale problems. It simultaneously seeks two decision hyperplanes, such that each hyperplane is required to be close to one of the two classes by the square loss function, and is required to be at least one distance apart from the other by hinge loss function. So it significantly reduces the computational time. Afterward, many researchers have made further improvements to TWSVM \cite{tbsvm}.
 	
As a successful improvement, nonparallel support vector machine \cite{npsvm} proposed by Tian et al. has become one of the state-of-the-art classifiers due to its great generalization performance. This model similarly seeks two nonparallel decision hyperplanes, and the hinge loss is also employed to allow the hyperplane as far as possible from other class. Unlike TWSVM, $\epsilon$-insensitive loss \cite{svr} replaces the original square loss to require that the hyperplane be as close to the class itself. It should be pointed that, TWSVM loses half of the sparsity due to the fact that the samples constrained by the square loss function almost all contribute to the final decision hyperplane. By contrast, the $\epsilon$-insensitive loss function is similar to the hinge loss function in that both allow only a fraction of the samples to be support vectors (the samples that contribute to the decision hyperplane). The $\epsilon$-insensitive loss gives the model the following merits: (a) Matrix inversion operation is avoided in the solving process. (b) Kernel trick can be implemented directly in the nonlinear case. (c) It follows the structural risk minimization (SRM) principle. (d) The sparsity of the model is improved. In this paper, the sparse property of NPSVM is denoted as whole sparsity and the corresponding property of TWSVM is denoted as semi-sparsity. In recent years, due to these advantages of the NPSVM, it has been combined with other learning theories to tackle different problems, such as multi-instance learning \cite{mit}, multi-view learning \cite{mvl}, multi-class learning \cite{mcl}, large margin distribution machine \cite{ldm}. These methods have all yielded excellent performances. So it is potentially beneficial to extend the NPSVM to handle multi-task issues.

For decades, multi-task learning (MTL) as a branch of machine learning, has developed rapidly in web application \cite{web}, bioinformatics \cite{bio}, computer vision \cite{cv}, and natural language processing \cite{nlp}. Compared with the STL methods, it improves the generalization performance via discovering relations among tasks, and supposes all related tasks have potential similar structural information\cite{infor}. Multi-task learning theory has thus been further supplemented and enhanced rapidly \cite{pro1,pro2}. Generally speaking, the MTL methods can be divided into three categories based on the content of the shared information, such as feature-based \cite{feature1,feature2}, instance-based \cite{instance} and parameter-based \cite{parame1,parame2} methods. The feature-based MTL assumes that multiple tasks share the same feature subspace and requires that the feature coefficients of multiple tasks are sparse. Instance-based MTL attempts to identify samples in each task that may be beneficial to other tasks. Parameter-based MTL assumes that multiple related tasks have common parameters.

Recently, the mean regularized multi-task learning (RMTL) \cite{rmtl} proposed by Evgeniou et al. firstly combines multi-task learning theory and support vector machine, and achieves good generalization performance. As a parameter-based MTL approach, RMTL assumes that all tasks share a common mean hyperplane, and the hyperplane of each task has an offset with the mean hyperplane. The final decision hyperplane of each task is determined by the common hyperplane and its offset. Whereas RMTL has a low computational efficiency due to the necessary to handle a large scale problem, by combining TWSVM with MTL, a direct multi-task twin support vector machine (DMTSVM) is further proposed by Xie et al. \cite{dmtsvm}. It simultaneously seeks two decision hyperplanes for each task, theoretically increasing computational efficiency by four times.

Due to the excellent performance of DMTSVM, many researchers have made many improvements. Multi-task centroid twin support vector machine (MTCTSVM) \cite{mtctsvm} proposed by Xie et al. additionally takes into account the centroid of each task. Mei et al. presented multi-task $v$-twin support vector machine (MT-$v$-TWSVM) \cite{vtwin} based on the property of $v$ in $v$-TWSVM, where the value of $v$ can control the sparsity of the model. Moreover, based on the idea that misclassified samples should be given different penalties in different locations, An et al. introduced rough set theory into MT-$v$-TWSVM and established a rough margin-based multi-task $v$-twin support vector machine (rough MT-$v$-TSVM) \cite{roughv}. The above multi-task TWSVMs all obtain better generalization performance due to their own unique structures, but they all have to face the following problems:
\begin{itemize}
 	\item [\textbullet]When processing these models, the matrix inversion operation is required. However, when the matrix is not invertible, the added correction term makes the result of the solution is not exactly equal to the optimal solution of the original model.
 	\item [\textbullet]These models need to consider extra kernel-generated space when using kernel trick \cite{kernel1} to solve linear inseparable problem. This increases the burden of model implementation.
\end{itemize}

Based on the ideas above, this paper puts forward a novel multi-task nonparallel support vector machine, it firstly introduces the idea of nonparallel support vector machine into the multi-task learning field. By replacing the square loss in the multi-task TWSVMs with $\epsilon$-insensitive loss, MTNPSVM not only considers the correlation between tasks when training multiple related tasks, but also inherits the merits of NPSVM. But it inevitably increases the scale of the problem. To address this problem, the ADMM \cite{admm} is adopted to accelerate computational efficiency by converting a large problem into multiple small problems. The main contributions of the paper can be summarized as follows:
\begin{enumerate}
\item This paper proposes a novel multi-task nonparallel support vector machine, which improves the generalization performance by introducing the $\epsilon$-insensitive loss function.
\item MTNPSVM constrains one class of samples by the $\epsilon$-insensitive loss instead of the square loss. This makes the samples appear only in the constraints, thus avoiding the matrix inversion operation and directly applying the kernel trick in the nonlinear case.
\item ADMM is employed in the MTNPSVM, which greatly improves the solving efficiency.
\end{enumerate}

The rest of this paper is outlined as follows. In Section \ref{10044}, a brief review of the DMTSVM and NPSVM is shown. MTNPSVM is proposed in Section \ref{10055}. A detailed derivation of ADMM to solve MTNPSVM is provided in Section \ref{10066}. A large number of comparative experiments have been shown in Section \ref{10077}. Finally, some conclusions and future directions for research are given in Section \ref{10088}.
\section{Related work}\label{10044}
In this section, detailed explanations of the nonparallel support vector machine and the direct multi-task support vector machine are shown, and these models are the basis of MTNPSVM.
\subsection{Nonparallel support vector machine}
As a single-task learning method, NPSVM is similar to TWSVM, which seeks two nonparallel proximal hyperplanes $x^{\top}w_{+}+b_{+}=0$ and $x^{\top}w_{-}+b_{-}=0$. Unlike TWSVM, the regularization term and the $\epsilon$-insensitive loss function are introduced into the model. The matrices $A_{+}$ and $B_{-}$ are defined as all positive and negative samples, respectively. For simplicity, the $A=(A_{+},e_{+})$, $B=(B_{-},e_{-})$, $u=(w_{+};b_{+})$, and $v=(v_{-};b_{-})$ are denoted, where $e_{+}$ and $e_{-}$ are vectors of ones of appropriate dimensions. Then the original problems of NPSVM are displayed as follows:
\begin{eqnarray}\label{1001}
\displaystyle{\min_{u,\xi,\xi^{*},\eta}}~~&&\frac{1}{2}\|u\|^{2}+C_{1}e_{+}^{\top}\left(\xi+\xi^{*}\right)+C_{2}e_{-}^{\top}\eta\\
\mbox{s.t.}~~&&-\varepsilon e_{+}-\xi^{*}\le\phi(A)u\le\varepsilon e_{+}+\xi, \nonumber\\
&&-\phi(B)u\ge e_{-}-\eta,\nonumber\\
&&\xi,~\xi^{*},~\eta\ge0,\nonumber
\end{eqnarray}
and
\begin{eqnarray}\label{1002}
\displaystyle{\min_{v,\xi,\eta,\eta^{*}}}~~&&\frac{1}{2}\|v\|^{2}+C_{3}e_{-}^{\top}\left(\eta+\eta^{*}\right)+C_{4}e_{+}^{\top}\xi\\
\mbox{s.t.}~~&&-\varepsilon e_{-}-\eta^{*}\le\phi(B)v\le\varepsilon e_{-}+\eta,\nonumber\\
&&\phi(A)v\ge e_{+}-\xi,\nonumber\\
&&\eta,~\eta^{*},~\xi\ge0\nonumber,
\end{eqnarray}
where $C_{i}\ge0$, $(i=1, 2, 3, 4)$ are trade-off parameters, $\xi$, $\xi^{*}$, $\eta$ and $\eta^{*}$ are slack variables. $\phi(\cdot)$ is the mapping function which can map the samples from the original space to the higher dimensional space, and the different nonlinear mapping can be exploited. In the linear case, the mapping function will degenerate into identity mapping.

As is shown in primal problem (\ref{1001}), when constructing positive hyperplane, $\epsilon$-insensitive loss function can restrict the positive samples in $\epsilon$-band between $x^{\top}w_{+}+b_{+}=\epsilon$ and $x^{\top}w_{+}+b_{+}=-\epsilon$ as much as possible. The hinge loss can make the negative samples at least 1 away from the positive hyperplane. This leaves the positive hyperplane determined by only a small number of samples in two classes. Thus, the $\epsilon$-insensitive loss function improves the model from semi-sparsity to whole sparsity. Moreover, the regularization term $\frac{1}{2}\|u\|^{2}$ is added to make the width of the $\epsilon$-band as large as possible, thus enabling the model to follow the SRM principle. In addition, this model avoids matrix inversion operation in the solving process. The same derivation happens in problem (\ref{1002}).

The dual formulations of problems (\ref{1001}) and (\ref{1002}) can be converted to the following form:
\begin{eqnarray}\label{1003}
\displaystyle{\min_{\pi}}~~&&\frac{1}{2}{\pi}^{\top}{\Lambda}{\pi}+{\kappa}^{\top}{\pi}\\
\text{s.t.}~~&&{e}^{\top}{\pi}=0,\nonumber\\
&&{0}\leq{\pi}\leq{C},\nonumber\
\end{eqnarray}
where $\Lambda$ is a matrix of appropriate size. $\pi$, $e$, $\kappa$ and $C$ are vectors of appropriate dimensions. It is observed that this form is a standard QPP, so the NPSVM can be solved efficiently by sequential minimization optimization (SMO) method or alternating direction method of multipliers (ADMM). Due to these incomparable advantages, the model performs better than other algorithms, but this method can only learn the tasks individually which is not favorable for learning multiple associated tasks.
\subsection{Direct multi-task twin support vector machine}\label{10033}
DMTSVM is built on the foundation of RMTL, which directly integrates the thoughts of TWSVM and MTL. In contrast to RMTL, this model constructs two nonparallel hyperplanes for each task, which reduces the scale of the problem and improves efficiency. Suppose $X_{p}$ and $X_{q}$ represent positive and negative samples of all tasks, respectively. $X_{pt}$ and $X_{qt}$ represent the positive and negative samples in the $t$-th task. $e_{t}$, $e_{1t}$, $e_{2t}$ and $e$ are one vectors of appropriate dimensions, the length of $e_{1t}$, $e_{2t}$ is equal to the number of positive and negative samples of the $t$-th task, respectively. The $A$=($X_{p}$, $e$), $B$=($X_{q}$, $e$), $A_{t}$=($X_{pt}$, $e_{1t}$) and $B_{t}$=($X_{qt}$, $e_{2t}$) are denoted. Based on the idea of multi-task learning, all tasks share two common hyperplanes $u$=$(w_{1}; b_{1})$ and $v$=$(w_{2}; b_{2})$. $u_{t}$ and $v_{t}$ represent the biases of $t$-task, respectively. The positive decision hyperplane of the $t$-th task can be expressed as ($w_{1t}$; $b_{1t}$)=($u$+$u_{t}$), while the negative decision hyperplane is ($w_{2t}$; $b_{2t}$)=($v$+$v_{t}$). DMTSVM is acquired by solving the following two QPPs:
\begin{eqnarray}\label{1004}
\displaystyle{\min_{u, u_{t}, p_{t}}}~~&&\frac{1}{2}\|Au\|_{2}^{2}+\frac{1}{2}\sum_{t=1}^{T}\rho_{t}\left\|A_{t}u_{t}\right\|_{2}^{2}+C_{1}\sum_{t=1}^{T}e_{2t}^{\top} \xi_{t}\\
\mbox{s.t.}~~&&-B_{t}\left(u+u_{t}\right)+\xi_{t}\geq e_{2t},\nonumber\\
~~&&\xi_{t}\geq0,~t=1,2,\cdots,T,\nonumber
\end{eqnarray}
and
\begin{eqnarray}\label{1005}
\displaystyle{\min_{v, v_{t}, q_{t}}}~~&&\frac{1}{2}\|Bv\|_{2}^{2}+\frac{1}{2}\sum_{t=1}^{T}\lambda_{t}\left\|B_{t}v_{t}\right\|_{2}^{2}+C_{2}\sum_{t=1}^{T} e_{1t}^{\top}\eta_{t} \\
\mbox{s.t.}~~&&A_{t}\left(v+v_{t}\right)+\eta_{t}\geq e_{1t},\nonumber\\
~~&&\eta_{t}\geq0,~t=1,2,\cdots,T,\nonumber
\end{eqnarray}
where $C_{i}\ge0, (i=1, 2)$ are trade-off parameters. $\xi_{t}$ and $\eta_{t}$ represent slack variables. $\rho_{t}$ and $\lambda_{t}$ can adjust the relationship between tasks. For the primal problem (\ref{1004}), when constructing the positive hyperplane for each task, the square loss in the objective function can restrict the hyperplane locate as close as possible to all positive samples, and the hinge loss can make the hyperplane be at least 1 away from the negative samples. A similar derivation occurs in problem (\ref{1005}). When $\rho_{t}\rightarrow$0 and $\lambda_{t}\rightarrow$0, this causes $u\rightarrow$0, $v\rightarrow$0 and all tasks are treated as unrelated. In contrary, when $\rho_{t}\rightarrow\infty$ and $\lambda_{t}\rightarrow\infty$, it leads to $u_{t}\rightarrow$0 and $v_{t}\rightarrow$0 and all tasks will considered as a unified whole. The label of $x$ in $t$-th task is assigned with the following decision function:
\begin{eqnarray}\label{1006}
f(x)=\arg\min_{r=1,2}\left|x^{\top}w_{rt}+b_{rt}\right|.
\end{eqnarray}

As an extension of TWSVM to multi-task learning scenario, DMTSVM can take advantage of correlation between tasks to improve generalization performance. However, this model has similar disadvantages to TWSVM, such that the semi-sparsity of the model, and the matrix inversion operation that cannot be avoided in the solving process.
\section{Multi-task nonparallel support vector machine}\label{10055}
In Section \ref{10044}, NPSVM and DMTSVM are proved to be complementary, so based on the above two models, a novel multi-task nonparallel support vector machine (MTNPSVM) is presented, it absorbs the merits of NPSVM and multi-task learning. This provides a modern perspective on the extension of NPSVM to multi-task learning.
\subsection{Linear MTNPSVM}
In this subsection, the definitions of matrices $A$, $B$, $A_{t}$, $B_{t}$ and the vectors $u$, $v$, $u_{t}$, $v_{t}$, $e_{1t}$, $e_{2t}$ are the identical to those utilized in section \ref{10033}. Also $u+u_{t}=(w_{1t}; b_{1t})$, $v$+$v_{t}=(w_{2t}; b_{2t})$ are vectors of positive plane and negative plane in the $t$-th task. The primal problems of MTNPSVM can be built as follows:
\begin{eqnarray}\label{1009}
\displaystyle{\min_{{u},u_{t},\eta_{t}^{*},\eta_{t}^{*},\xi_{t}}}~~&&\frac{\rho_{1}}{2}\|u\|^{2}+\frac{1}{2T}\sum_{t=1}^{T}\left\|u_{t}\right\|^{2}+C_{1}\sum_{t=1}^{T} e_{1t}^{\top}\left(\eta_{t}+\eta_{t}^{*}\right)+C_{2}\sum_{t=1}^{T}e_{2t}^{\top}\xi_{t}\\
\mbox{s.t.~~~}~~&&-\epsilon e_{1t}-\eta_{t}^{*}\leq A_{t}\left(u+u_{t}\right)\leq\epsilon{e}_{1 t}+\eta_{t},\nonumber\\
&&B_{t}\left(u+u_{t}\right)\leq-e_{2t}+\xi_{t},\nonumber\\
&&\eta_{t},~\eta_{t}^{*},~\xi_{t}\geq0,~t=1,2,\cdots,T,\nonumber
\end{eqnarray}
and
\begin{eqnarray}\label{1010}
\displaystyle{\min _{v, v_{t},\xi_{t}^{*},\xi_{t}^{*},\eta_{t}}}~~&&\frac{\rho_{2}}{2}\|v\|^{2}+\frac{1}{2T}\sum_{t=1}^{T}\left\|v_{t}\right\|^{2}+C_{3}\sum_{t=1}^{T} e_{2t}^{\top}\left(\xi_{t}+\xi_{t}^{*}\right)+C_{4}\sum_{t=1}^{T}e_{1t}^{\top}\eta_{t}\\
\mbox{s.t.~~~}~~&&-\epsilon e_{2t}-\xi_{t}^{*}\leq B_{t}\left(v+v_{t}\right)\leq\epsilon{e}_{2t}+\xi_{t},\nonumber\\
&&A_{t}\left(v+v_{t}\right)\geq e_{1t}-\eta_{t},\nonumber\\
&&\xi_{t},~\xi_{t}^{*},~\eta_{t}\geq0,~t=1,2,\cdots,T.\nonumber
\end{eqnarray}

The relationship between tasks can be adjusted by $\rho_{1}$ and $\rho_{2}$. $C_{i}\ge 0$, ($i$=1, 2, $\cdots$, 4) are penalty parameters. $\xi_{t}$, $\xi_{t}^{*}$, $\eta_{t}$ and $\eta_{t}^{*}$ are slack variables of the $t$-th task like the corresponding parameters in NPSVM.

Note that the primal problem (\ref{1009}), when constructing the positive hyperplane for each task, the $\epsilon$-insensitive loss $\left(\eta_{t}+\eta_{t}^{*}\right)$ accompanied by the first constraint can restrict the positive samples in $\epsilon$-band between $x^{\top}w_{1t}+b_{1t}=\epsilon$ and $x^{\top}w_{1t}+b_{1t}=-\epsilon$ as much as possible, and the hinge loss $\xi_{t}$ accompanied by the second constraint can allow the hyperplane be at least 1 away from the negative samples. In addition, MTNPSVM can obtain the commonality between tasks through the parameter $u(v)$ and capture the personality of each task through the parameter $u_{t}(v_{t})$. Also the first two regularization terms are equivalent to the trade-off between maximizing the width of $\epsilon$-band $\frac{2\epsilon}{\|w_{1t}\|}$ and minimizing the distance between each task hyperplane and the common hyperplane. The similar conclusion can be found in \cite{npsvm,rmtl}. The construction of the negative hyperplane in problem (\ref{1010}) is similar to that in problem (\ref{1009}).

The dual problems of (\ref{1009}) and (\ref{1010}) can be obtained by introducing the Lagrangian multiplier vectors $\alpha_{t}^{+}$, $\alpha_{t}^{+*}$, $\beta_{t}^{-}$, $\gamma_{t}$, $\theta_{t}$, $\psi_{t}$. Now taking the problem (\ref{1009}) as an example. The Lagrangian function can be given by
\begin{eqnarray}\label{20001}
L=&&\frac{\rho_{1}}{2}\|u\|^{2}+\frac{1}{2T}\sum_{t=1}^{T}\left\|u_{t}\right\|^{2}+C_{1} \sum_{t=1}^{T} e_{1t}^{\top}\left(\eta_{t}+\eta_{t}^{*}\right)+C_{2} \sum_{t=1}^{T}e_{2 t}^{\top}\xi_{t}\nonumber\\&&-\sum_{t=1}^{T}\alpha_{t}^{+\top}\left[\epsilon e_{1t}+\eta_{t}-A_{t}\left(u+u_{t}\right)\right]-\sum_{t=1}^{T}\alpha_{t}^{+*\top}\left[\epsilon e_{1t}+\eta_{t}^{*}+A_{t}\left(u+u_{t}\right)\right]\nonumber\\
&&-\sum_{t=1}^{T}\beta_{t}^{-\top}\left[-e_{2t}+\xi_{t}-B_{t}\left(u+u_{t}\right)\right]-\sum_{t=1}^{T}\gamma_{t}^{\top}\xi_{t}-\sum_{t=1}^{T}\theta_{t}^{\top} \eta_{t}-\sum_{t=1}^{T}\psi_{t}^{\top}\eta_{t}^{*},
\end{eqnarray}
the KKT conditions can be obtained by differentiating parameters $u$, $u_{t}$, $\eta_{t}$, $\eta_{t}^{*}$, $\xi_{t}$ and setting the differential equations equal to 0:
\begin{eqnarray}
&&\frac{\partial L}{\partial u}=\rho_{1}u-\sum_{t=1}^{T}A_{t}^{\top}\left(\alpha_{t}^{+*}-\alpha_{t}^{+}\right)+\sum_{t=1}^{T}B_{t}^{\top}\beta_{t}^{-}=0, \\
&&\frac{\partial L}{\partial v_{t}}=\frac{u_{t}}{T}-A_{t}^{\top}\left(\alpha_{t}^{+*}-\alpha_{t}^{+}\right)+B_{t}^{\top}\beta_{t}^{-}=0, \\
&&\frac{\partial L}{\partial \eta_{t}}=C_{1}e_{1 t}-\alpha_{t}^{+}-\theta_{t}=0, \\
&&\frac{\partial L}{\partial \eta_{t}^{*}}=C_{1} e_{1 t}-\alpha_{t}^{+*}-\psi_{t}=0, \\
&&\frac{\partial L}{\partial \xi_{t}}=C_{2}e_{2 t}-\beta_{t}^{-}-\gamma_{t}=0.
\end{eqnarray}

By the above equations, the polynomial for each parameter can be derived, then substituting them into the original Lagrangian function. By declaring the following definition:
\begin{eqnarray}\label{30001}
&&P_{t}=A_{t}\cdot B_{t}^{\top},\\
&&P=blkdiag(P_{1},P_{2},\cdots,P_{T}),\\
&&M(A,B^{\top})=\frac{1}{\rho}A\cdot B^{\top}+T \cdot P,
\end{eqnarray}
where $blkdiag(\cdot)$ is used to construct the block diagonal matrix, the dual form can be given as follows:
\begin{eqnarray}\label{dual1}
\displaystyle{\min_{\alpha^{+*},\alpha^{+},\beta^{-}}}~~&&\frac{1}{2}\left(\alpha^{+*}-\alpha^{+}\right)^{\top} M(A,A^{\top})\left(\alpha^{+*}-\alpha^{+}\right)-\left(\alpha^{+*}-\alpha^{+}\right)^{\top}M(A,B^{\top})\beta^{-}\nonumber\\
&&+\frac{1}{2}\beta^{-\top}M(B,B^{\top})\beta^{-}+\epsilon e_{1}^{\top}\left(\alpha^{*}+\alpha\right)-e_{2}^{\top} \beta^{-} \\
\mbox {s.t.~~~}~~&&0\le\alpha^{+},~\alpha^{+*} \le C_{1} e_{1},\nonumber\\
~~&&0\le\beta^{-}\le C_{2}e_{2},\nonumber
\end{eqnarray}
where $\alpha^{+*}$=$(\alpha_{1}^{+*}; \cdots; \alpha_{t}^{+*})$, $\alpha^{+}$=$(\alpha_{1}^{+}; \cdots; \alpha_{t}^{+})$, and $\beta^{-}$=$(\beta_{1}^{-}; \cdots; \beta_{t}^{-})$. $e_{1}$ and $e_{2}$ are the ones vectors of approximate dimensions. By further simplifying the above equations, the dual formulation of problem (\ref{1009}) can be concisely rewritten as
\begin{eqnarray}\label{1012}
\displaystyle{\min_{\widetilde{\pi}}}~~&&\frac{1}{2} \widetilde{\pi}^{\top} \widetilde{\Lambda} \widetilde{\pi}+\widetilde{\kappa}^{\top} \widetilde{\pi} \\
\mbox {s.t.}~~&&0 \leq\widetilde{\pi} \leq \widetilde{C}.\nonumber\
\end{eqnarray}
Here $\widetilde{\Lambda}$=
$\begin{pmatrix}
H_{1} &-H_{2}\\
-\hat{H_{2}}&H_{3}
\end{pmatrix}$,
$H_{1}$=
$\begin{pmatrix}
M(A,A^{\top}) &-M(A,A^{\top})\\
-M(A,A^{\top})&M(A,A^{\top})
\end{pmatrix}$,
$H_{2}$=
$\begin{pmatrix}
M(A,B^{\top}) &\\
-M(A,B^{\top})&
\end{pmatrix}$,
$H_{3}$=M($B$, $B^{\top}$),
and $\widetilde{\pi}=(\alpha^{+*};\alpha^{+};\beta^{-})$,
$\widetilde{C}=(C_{1}e_{1};C_{1}e_{1};C_{2}e_{2})$, and
$\widetilde{\kappa}=(\epsilon e_{1};\epsilon e_{1};-e_{2})$.

The problem of (\ref{1012}) is clearly a QPP. Similarly the dual form of (\ref{1010}) is shown as follows:
\begin{eqnarray}
\label{dual2}
\displaystyle{\min_{\alpha^{-*},\alpha^{-},\beta^{+}}}~~&& \frac{1}{2}\left(\alpha^{-*}-\alpha^{-}\right)^{\top} M(B,B^{\top})\left(\alpha^{-*}-\alpha^{-}\right)-\left(\alpha^{-*}-\alpha^{-}\right)^{\top}M(B,A^{\top})\beta^{+}\nonumber\\
&&+\frac{1}{2}\beta^{+\top} M(A,A^{\top})\beta^{+}
+\epsilon e_{2}^{\top}\left(\alpha^{-*}+\alpha^{-}\right)-e_{1}^{\top}\beta^{+} \\
\mbox {s.t.~~~}~~&&0 \le\alpha^{-},\alpha^{-*}\le C_{3}e_{2}, \nonumber\\
&&0\le\beta^{+}\le C_{4}e_{1}.\nonumber
\end{eqnarray}
 Similarly, $\alpha^{-*}=(\alpha_{1}^{-*};\cdots;\alpha_{t}^{+*})$, $\alpha^{-}$=$(\alpha_{1}^{-};\cdots;\alpha_{t}^{-})$, $\beta^{+}$=$(\beta_{1}^{+};\cdots;\beta_{t}^{+})$, and the dual problem can be concisely reformulated as
 \begin{eqnarray}\label{1013}
 \displaystyle{\min_{\hat{\pi}}}~~&&\frac{1}{2} \hat{\pi}^{\top} \hat{\Lambda} \hat{\pi}+\hat{\kappa}^{\top} \hat{\pi} \\
 \mbox {s.t.}~~&& 0 \le \hat{\pi} \le \hat{C}.\nonumber\
 \end{eqnarray}
Here
$\hat{\Lambda}$=
$\begin{pmatrix}
Q_{1} &-Q_{2}\\
-Q_{2}&Q_{3}
\end{pmatrix}$,
$Q_{1}$=				
$\begin{pmatrix}
M(B,B^{\top}) &-M(B,B^{\top})\\
-M(B,B^{\top})&M(B,B^{\top})
\end{pmatrix}$,
$Q_{2}$=
$\begin{pmatrix}
M(B,A^{\top}) &\\
-M(B,A^{\top})&
\end{pmatrix}$,
$Q_{3}=M(A, A^{\top})$, $\hat{\pi}=(\alpha^{-*};\alpha^{-};\beta^{+})$,
$\hat{C}=(C_{3}e_{2};C_{3}e_{2};C_{4}e_{1})$, and
$\hat{\kappa}=(\epsilon e_{2};\epsilon e_{2}; -e_{1})$.

The following conclusions can be justified by applying the KKT conditions of problems (\ref{1012}) and (\ref{1013}). The proofs of Theorems \ref{th1} and \ref{th3} are placed in Appendix \ref{proof1}, and the proofs of Theorems \ref{th2} and \ref{th4} are shown in Appendix \ref{proof2}. The similar conclusion can also be found in \cite{npsvm,mvl}.
\begin{theorem}\label{th1}
Suppose $\widetilde{\pi}^{*}$ is the optimal solution of (\ref{1012}), if $\alpha_{it}^{+}$ and $\alpha_{it}^{+*}$ represent the $i$-th component of $\alpha_{t}^{+}$ and $\alpha_{t}^{+*}$, respectively. The each pair of $\alpha_{it}^{+*}$ and $\alpha_{it}^{+}$ must satisfy $\alpha_{it}^{+*}\alpha_{it}^{+}=0$, $i=1, 2,\cdots, q$; $t=1, 2,\cdots, T$, which implies that the each pair parameters can not be nonzero simultaneously.
\end{theorem}
\begin{theorem}\label{th2}
Suppose $\widetilde{\pi}^{*}$ is the optimal solution of (\ref{1012}), the value of $u$ can be obtained by applying the KKT conditions of (\ref{1009}) in the following way:
\begin{eqnarray}\label{h1}
u=\frac{1}{\rho_{1}}(\sum_{t=1}^{T} A_{t}^{\top}\left(\alpha_{t}^{+*}-\alpha_{t}^{+}\right)-\sum_{t=1}^{T} B_{t}^{\top} \beta_{t}^{-}),
\end{eqnarray}
\begin{eqnarray}\label{h2}
u_{t}=T(A_{t}^{\top}\left(\alpha_{t}^{+*}-\alpha_{t}^{+}\right)-B_{t}^{\top} \beta_{t}^{-}).
\end{eqnarray}
\end{theorem}
\begin{theorem}\label{th3}
Suppose $\hat{\pi}^{*}$ is the optimal solution of (\ref{1013}), if $\alpha_{it}^{-}$ and $\alpha_{it}^{-*}$ represent the $i$-th component of $\alpha_{t}^{-}$ and $\alpha_{t}^{-*}$, respectively. The each pair of $\alpha_{it}^{-*}$ and $\alpha_{it}^{-*}$ must satisfy $\alpha_{it}^{-*}$$\alpha_{it}^{-}=0$, $i=1, 2,\cdots, q$; $t=1, 2,\cdots, T$, which implies that the each pair parameters can not be nonzero simultaneously.
\end{theorem}
\begin{theorem}\label{th4}
Suppose $\hat{\pi}^{*}$ is the optimal solution of (\ref{1013}), the value of $u$ can be obtained by applying the KKT conditions of (\ref{1010}) in the following way:
\begin{eqnarray}\label{h3}
v=\frac{1}{\rho_{2}}(\sum_{t=1}^{T} B_{t}^{\top}\left(\alpha_{t}^{-*}-\alpha_{t}^{-}\right)+\sum_{t=1}^{T} A_{t}^{\top} \beta_{t}^{+}),
\end{eqnarray}
\begin{eqnarray}\label{h4}
v_{t}=T(B_{t}^{\top}\left(\alpha_{t}^{-*}-\alpha_{t}^{-}\right)-A_{t}^{\top} \beta_{t}^{+}).
\end{eqnarray}
\end{theorem}

In terms of Theorems \ref{th2} and \ref{th4}, there is no necessary to calculate the inversion matrix when obtaining the parameters of mean hyperplane and bias, which can accelerate the computational speed to a certain extent. Combined with the $u+u_{t}=(w_{1t}; b_{1t})$, $v$+$v_{t}=(w_{2t}; b_{2t})$, the label of the test sample $x$ in $t$-th task can obtained by the following equation:
\begin{eqnarray}\label{30003}
f(x)=\arg\min_{r=1,2} \left|x^{\top}w_{rt}+b_{rt}\right|.
\end{eqnarray}
\subsection{Nonlinear MTNPSVM}
Unlike the multi-task TWSVMs, MTNPSVM can directly exploit the kernel trick in the nonlinear case and thus only needs to deal with the problems similar to the linear case. The reason is that the nonlinear mapping function appears only as the inner product in the dual problem. $\phi(\cdot)$ represents the nonlinear mapping function, $x_{it}$ represents random sample. Finally, the decision hyperplanes of the $t$-th task will be changed as follows:
\begin{eqnarray}\label{2008}
\phi(x_{it})^{\top}w_{1t}+b_{1t}=0,\text{ and } \phi(x_{it})^{\top}w_{2t}+b_{2t}=0.
\end{eqnarray}
To obtain the above hyperplanes, the nonlinear MTNPSVM needs to solve the following problems:
\begin{eqnarray}\label{2009}
\displaystyle{\min_{u, u_{t},\eta_{t}^,\eta_{t}^{*}\,xi_{t}}}~~&&\frac{\rho_{1}}{2}\|u\|^{2}+\frac{1}{2T}\sum_{t=1}^{T}\left\|u_{t}\right\|^{2}+C_{1}\sum_{t=1}^{T} e_{1t}^{\top}\left(\eta_{t}+\eta_{t}^{*}\right)+C_{2}\sum_{t=1}^{T}e_{2t}^{\top}\xi_{t} \\
\mbox{s.t.}~~&&-\epsilon e_{1t}-\eta_{t}^{*} \leq\phi\left(A_{t}\right)\left(u+u_{t}\right)\leq\epsilon e_{1t}+\eta_{t},\nonumber\\
&&\phi\left(B_{t}\right)\left(u+u_{t}\right)\leq-e_{2t}+\xi_{t},\nonumber\\
&&\eta_{t},~\eta_{t}^{*},~\xi_{t}\geq 0,~t=1,2,\cdots,T,\nonumber
\end{eqnarray}
and
\begin{eqnarray}\label{2010}
\displaystyle{\min_{v, v_{t},\xi_{t},\xi_{t}^{*},\eta_{t}}}~~&&\frac{\rho_{2}}{2}\|v\|^{2}+\frac{1}{2T} \sum_{t=1}^{T}\left\|v_{t}\right\|^{2}+C_{3}\sum_{t=1}^{T}e_{2t}^{\top}\left(\xi_{t}+\xi_{t}^{*}\right)+C_{4}\sum_{t=1}^{T}e_{1 t}^{\top} \eta_{t}\\
\mbox{s.t.}~~&&-\epsilon e_{2t}-\xi_{t}^{*} \leq\phi\left(B_{t}\right)\left(v+v_{t}\right)\leq \epsilon e_{2t}+\xi_{t}, \nonumber\\
&&\phi\left(A_{t}\right) \left(v+v_{t}\right) \geq e_{1t}-\eta_{t},\nonumber\\
&&\xi_{t},~\xi_{t}^{*},~\eta_{t}\geq 0,~t=1,2,\cdots,T.\nonumber
\end{eqnarray}

The original problem is almost identical to the linear case, except that the mapping function $\phi$($\cdot$) is introduced into the primal problems. A corresponding difference in the dual problem is the definition of (\ref{30001}). In the nonlinear case, the new definition is as follows:
\begin{eqnarray}\label{30002}
&&P_{t}=K(A_{t}, B_{t}^{\top}),\\
&&P=blkdiag(P_{1},P_{2},\cdots,P_{T}),\\
&&M(A,B^{\top})=\frac{1}{\rho}K(A, B^{\top})+T \cdot P,
\end{eqnarray}
here $K(x_{i},x_{j})=(\phi(x_{i})\cdot\phi(x_{j}))$ represents kernel function, the Polynomial kernel and the RBF kernel are employed in this paper. The properties in the nonlinear case are very similar to Theorems \ref{th1}$\sim $\ref{th4}, this only requires transforming the identical mapping into the nonlinear mapping function. Finally, the label of a new sample can be obtained by the same decision function as (\ref{30003}).
\subsection{Advantages of MTNPSVM}
As an improvement of the DMTSVM, the MTNPSVM draws on the advantages of the NPSVM and avoids many disadvantages of the DMTSVM, thus it has significant theoretical merits. Although MTNPVM have a additional parameter $\epsilon$, it still has the following advantages:
\begin{itemize}
	\item[\textbullet] MTNPSVM has a similar elegant equation form as RMTL, which can avoid matrix inversion operation in the solving process. Moreover, it can be solved by SMO-type algorithms.
	\item[\textbullet] Only the inner product appears in the dual problem leading to the kernel trick can be directly employed in the nonlinear case. This reduces the burden on the implementation methods.
    \item[\textbullet] The inclusion of two regularization terms allows the model to reflect the commonality and individuality of tasks when dealing with multiple associated tasks. Also like RMTL, this enables the model to comply with the SRM principle.
	\item[\textbullet] DMTSVM loses sparsity due to the square loss function. In the proposed model MTNPSVM, the $\epsilon$-insensitive loss function is added so that it inherits the whole sparsity of the NPSVM. Models with high sparsity can be combined with algorithms, such as safe screening rule \cite{ssrnp,ssrdmt}, to speed up the efficiency of model solving.
\end{itemize}
\section{ADMM Optimization}\label{10066}
\subsection{ADMM for MTNPSVM}
MTNPSVM has a low efficiency in solving process due to the construction of large-scale matrices in the MTL methods. So the ADMM algorithm is developed into multi-task learning to accelerate the solving of MTNPSVM. ADMM is an advanced fast solving algorithm which improves computational efficiency by transforming a large scale problem into multiple small subproblems. In order to apply this algorithm, the inequality constraints of problems (\ref{1012}) and (\ref{1013}) are turned into the equality constraints. In this subsection, the details of solving MTNPSVM are displayed. By introducing new variables $\widetilde\lambda$ and $\hat{\lambda}$, the problems can be written as:
\begin{eqnarray}\label{1018}
\displaystyle{\min_{\widetilde{\pi}}}~~&&\frac{1}{2}\widetilde{\pi}^{\top}\widetilde{\Lambda}\widetilde{\pi}+\widetilde{\kappa}^{\top} \widetilde{\pi}+g(\widetilde\lambda)\\
\mbox{s.t.}~~&&\widetilde{\pi}+\widetilde\lambda=\widetilde{C},\nonumber
\end{eqnarray}
and
\begin{eqnarray}\label{1019}
\displaystyle{\min_{\hat{\pi}}}~~&&\frac{1}{2}\hat{\pi}^{\top}\hat{\Lambda}\hat{\pi}+\hat{\kappa}^{\top}\hat{\pi}+g(\hat\lambda)\\
\mbox{s.t.}~~&&\hat{\pi}+\hat\lambda=\hat{C},\nonumber
\end{eqnarray}
where g($\cdot$) stands for indicator function, it is defined as (\ref{2001}), the value of the parameter $C$ changes according to the different functions.
\begin{eqnarray}\label{2001}
g(\lambda)=\left\{\begin{array}{ll}0, &\mbox {if}~0\le\lambda \le C \\+\infty, &\mbox {otherwise.}\end{array}\right.
\end{eqnarray}
 Then, the iterative procedures of ADMM algorithm for (\ref{1018}) and (\ref{1019}) is displayed as:
 \begin{eqnarray}\label{36}
\left\{\begin{aligned}
\widetilde{\pi}_{k+1}=&\mathop{\arg\min}\limits_{\pi}(\frac{1}{2}\widetilde{\pi}^{\top}\widetilde{\Lambda} \widetilde{\pi}+\widetilde{{\kappa}}^{\top}\widetilde{\pi}+\frac{\mu}{2}\|\widetilde{\pi}+\widetilde\lambda_{k}-\widetilde{C}+\widetilde{h}_{k}\|^{2}),\\
\widetilde{\lambda}_{k+1}=&\mathop{\arg\min}\limits_{\lambda}(g(\widetilde{\lambda})+\frac{\mu}{2}\|\widetilde{\pi}_{k+1}+\widetilde\lambda-\widetilde{C}+\widetilde{h}_{k}\|^{2}),\\
\widetilde{h}_{k+1}=&\widetilde{\pi}_{k+1}+\widetilde\lambda_{k+1}-\widetilde{C}+\widetilde{h}_{k},
\end{aligned}\right.
\end{eqnarray}
and
\begin{eqnarray}\label{37}
\left\{\begin{aligned}
\hat{\pi}_{k+1}=&\mathop{\arg\min}\limits_{\pi}(\frac{1}{2}\hat{\pi}^{\top}\hat{\Lambda} \hat{\pi}+\hat{{\kappa}}^{\top}\hat{\pi}+\frac{\mu}{2}\|\hat{\pi}+\hat\lambda_{k}-\hat{C}+\hat{h}_{k}\|^{2}),\\	\hat{\lambda}_{k+1}=&\mathop{\arg\min}\limits_{\lambda}(g(\hat{\lambda})+\frac{\mu}{2}\|\hat{\pi}_{k+1}+\hat\lambda-\hat{C}+\hat{h}_{k}\|^{2}),\\
\hat{h}_{k+1}=&\hat{\pi}_{k+1}+\hat\lambda_{k+1}-\hat{c}+\hat{h}_{k}.
\end{aligned}\right.
\end{eqnarray}
Here $k$ stands for the $k$-th iteration and $\mu$ is a relaxation factor which can control the speed of convergence. In algorithms, $f$ is denoted as the objective function value, the primal residual $r^{k+1}=\pi_{k+1}-\lambda_{k+1}$, the dual residual $s^{k+1}=\mu(\lambda_{k+1}-\lambda_{k})$. The convergence thresholds $\delta_{p}^{k}$, $\delta_{d}^{k}$ both are defined as the linear combination of the absolute tolerance $\delta_{1}$ and the relative tolerance $\delta_{2}$ as follows:
\begin{eqnarray}
	\label{x1}&&\delta_{p}^{k}=\delta_{1}\cdot\sqrt{n}+\delta_{2}\cdot\max (\|\pi_{k}\|,\|\lambda_{k}\|),\\
	\label{x2}&&\delta_{d}^{k}=\delta_{1}\cdot\sqrt{n}+\delta_{2}\cdot\|\mu h_{k}\|,
\end{eqnarray}where $n$ is the dimension of the vector $\pi_k$. If $\|r^{k}\|\le\delta_{p}^{k}$ and $\|s^{k}\|\le\delta_{d}^{k}$, the iteration will stop and the objective function value $f$ will converge to the certain value. The detailed derivation of the algorithm can be found in \cite{admm}.

Furthermore, the linear case is used as an instance to elaborate the overall process of algorithm optimization. Before solving, the original dual problems (\ref{1012}) and (\ref{1013}) must be transformed into the objective functions (\ref{1018}) and (\ref{1019}), which are the standard form of the objective function of the ADMM algorithm. The pseudo-code for solving the objective functions (\ref{1018}) and (\ref{1019}) is summarized in Algorithms. \ref{alg1} and \ref{alg2}, respectively.
	
\begin{figure}[!htbp]	
	\centering	
	\includegraphics[width=0.8\textwidth,height=0.3\textheight]{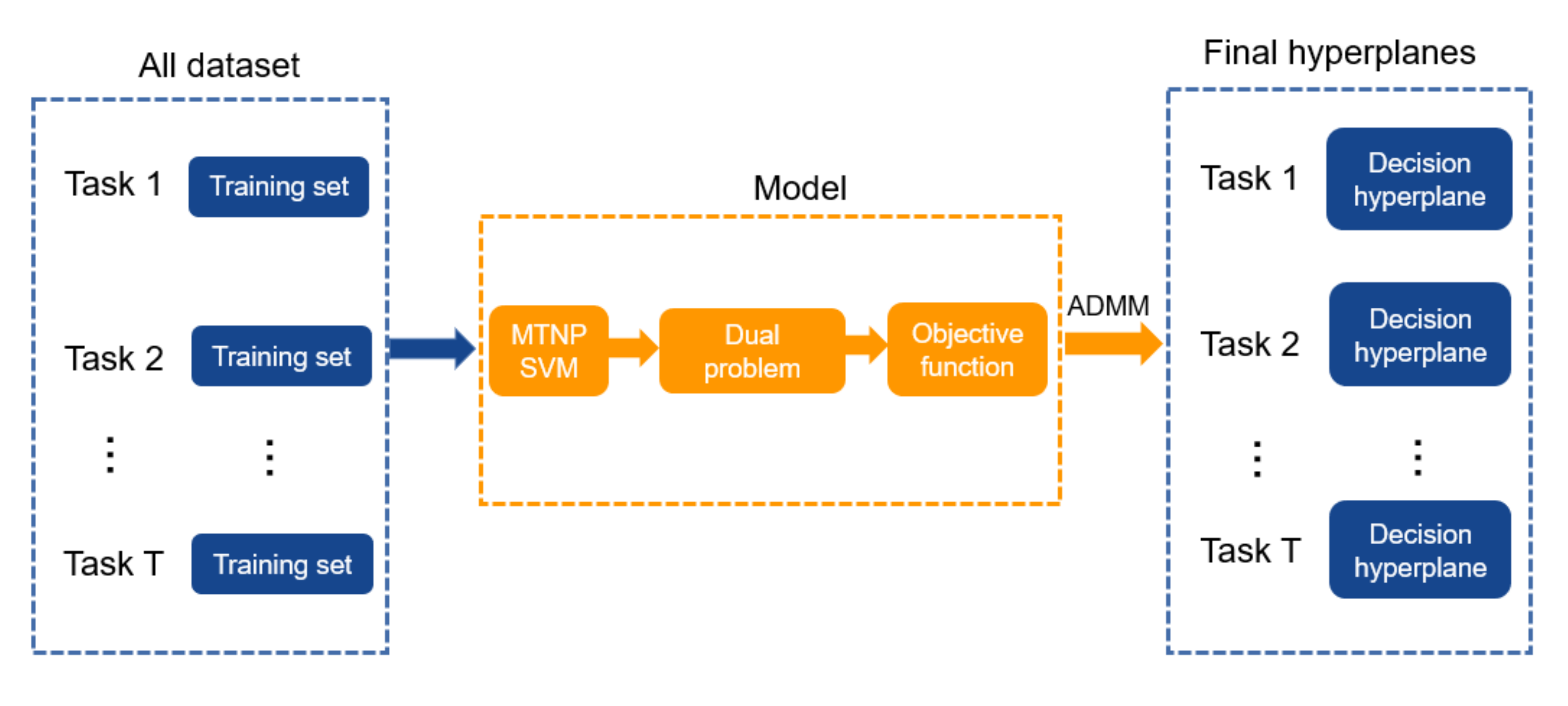}
	\caption{A description of the model construction and solution process. The datasets for all tasks are correlated and the decision hyperplane for each task can be obtained simultaneously by solving the proposed MTNPSVM.}
	\label{figx}
\end{figure}

Above all, the solving process of MTNPSVM are shown in Fig. \ref{figx}. As shown, MTNPSVM follows the classical multi-task learning framework. It is worth noting that the model needs to be transformed twice into the objective function of ADMM algorithm.
\begin{algorithm}[h]
	\caption{ADMM algorithm for objective function (\ref{1018})}
	\label{alg1}
	\begin{algorithmic}
    	\Require Train set $\{X,Y\}$ for all tasks; $A$, $B$ represent the positive and negative samples in the train set $X$; parameters $C_{1}$, $C_{2}$, $\epsilon$;
		\Ensure $u$, $u_{t}$;
		\State Initialize $\widetilde{\pi}_{0}$=$(\alpha^{+*};\alpha^{+};\beta^{-})$=0, $\widetilde{\lambda}_{0}$, $\widetilde{h}_{0}$, $\mu$, $k$=0;
		\State Compute $\widetilde{\Lambda}$ matrix according to the description of (\ref{1012});
		\Repeat
		\State Compute ($\widetilde{\pi}_{k+1}$, $\widetilde{\lambda}_{k+1}$, $\widetilde{h}_{k+1}$) by (\ref{36});
		\State Compute convergence thresholds $\delta_{p}^{k}$, $\delta_{d}^{k}$ via (\ref{x1}) and (\ref{x2});
		\Until $\|r^{k}\|\le\delta_{p}^{k}$, $\|s^{k}\|\le\delta_{d}^{k}$.
		\State Get the optimal solution $\widetilde{\pi}^{*}$=$\widetilde{\pi}_{k+1}$;
		\State Compute $u$, $u_{t}$ by (\ref{h1}) and (\ref{h2}).
	\end{algorithmic}
\end{algorithm}
\begin{algorithm}
	\caption{ADMM algorithm for objective function (\ref{1019})}
	\label{alg2}
	\begin{algorithmic}
		\Require Train set $\{X,Y\}$ for all tasks; $A$, $B$ represent the positive and negative samples in the train set $X$; parameters $C_{3}$, $C_{4}$, $\epsilon$;
		\Ensure $v$, $v_{t}$;
		\State Initialize $\hat{\pi}_{0}$=$(\alpha^{-*};\alpha^{-};\beta^{+})=0$, $\hat{\lambda}_{0}$, $\hat{h}_{0}$, $\mu$, $k$=0;
		\State Compute $\hat{\Lambda}$ matrix according to the description of (\ref{1013});
		\Repeat
		\State Compute ($\widetilde{\pi}_{k+1}$, $\hat{\lambda}_{k+1}$, $\hat{h}_{k+1}$) by (\ref{37});
		\State Compute convergence thresholds $\delta_{p}^{k}$, $\delta_{d}^{k}$ via (\ref{x1}) and (\ref{x2});
		\Until $\|r^{k}\|\le\delta_{p}^{k}$, $\|s^{k}\|\le\delta_{d}^{k}$;
		\State Get the optimal solution $\hat{\pi}^{*}$=$\hat{\pi}_{k+1}$;
		\State Compute $v$, $v_{t}$ by (\ref{h3}) and (\ref{h4}).
	\end{algorithmic}
\end{algorithm}
\subsection{Computational complexity}
This subsection theoretically analyzes the time complexity of algorithm. $p$, $q$ represent the number of positive and negative samples, respectively. Algorithm \ref{alg1} is used here as an example. The dimension of matrix $\widetilde{\Lambda}$ is calculated as $(2p+q)*(2p+q)$. $r$ represents the number of iterations. When updating the $\widetilde{\pi}$, since it needs to use the Choleskey decomposition in the first iteration, and store for subsequent calculations, so the computational complexity is O($(2p+q)^{3}$+r$(2p+q)^{2}$).
When updating the $\widetilde{\lambda}$ and $\widetilde{h}$, their computational complexities are all O($r(2p+q)^{2}$). The total computational complexity of ADMM algorithm is O($(2p+q)^{3}$+$r(2p+q)^{2}$). Also if the function ``quadprog" in MATLAB is used to solve it, the computational complexity is O($r(2p+q)^{3}$).

Apparently, if the number of iterations is exceeds $1$ and equal, the ADMM algorithm will have a theoretical advantage of higher computational efficiency. To verify the advantage of the convergence speed of the ADMM algorithm, the solving speeds of the ADMM and the ``quadprog" function further are compared practically in Section \ref{sec_con}.
\section{Numerical experiments}\label{10077}
In this section, three types of experiments including fifteen benchmark datasets, twelve image datasets, and one real Chinese Wine dataset are conducted to verify the validity of proposed MTNPSVM. Additional three benchmark datasets, often used to evaluate MTL methods, are utilized to analyze the properties of the model in Section \ref{analysis}. The STL methods used in experiments include SVM, TWSVM, $\nu$-TWSVM, LSTSVM and NPSVM, the MTL methods consist of DMTSVM, MTPSVM \cite{mtpsvm}, MTLS-TSVM \cite{mtlstsvm}, and MTCTSVM. Two kernels including RBF kernel and Polynomial kernel are used in the following experiments. Kernel parameter $\delta$ in RBF varies in $\{2^{i}|i=-3,-2,\cdots,3\}$, and Kernel parameter $\delta$ in Polynomial is selected in the set $\{1, 2, \cdots, 7\}$. Parameter $\epsilon$ in every model varies in $\{0.1, 0.2 \cdots, 0.5\}$. Parameter $\nu$ in $\nu$-TWSVM are selected in set $\{0.1, 0.2, \cdots,0.9\}$. The other parameters used in the experiment are chosen from the set $\{2^{i}|i=-3,-2,\cdots,3\}$. In addition, the``Accuracy" in experiments represents the mean accuracy of $T$ tasks. It can be calculated as $Accuracy= 1/T\sum_{t=1}^{T}Acc_{t}$, where the $Acc_{t}$ indicates the accuracy of $t$-th task. The ``time" denotes the computational cost for training datasets. For the comparability of experimental results, all the experiments are performed on Windows 10 running in MATLAB R2018b with the system configuration of Intel(R) Core(TM) I5-7200U CPU (2.50 GHz) with 8.00 GB of RAM. The code for the experiment can be downloaded from the web page\footnote{https://www.github.com/liuzongmin/mtnpsvm}.
\subsection{Experiments on fifteen benchmark datasets}
In this subsection, the performance of the MTNPSVM is demonstrated by conducting fifteen benchmark experiments with the seven methods. Here the methods contain two STL methods which are TWSVM and NPSVM, and five MTL methods consist of DMTSVM, MTPSVM, MTLS-TSVM, MTCTSVM and MTNPSVM. Each experimental dataset is divided into 80\% training set and 20\% testing set. The grid-search strategy and 5-fold cross-validation are performed in training set. More specially, training set is randomly divided into five subsets, one of which is used as the validation set and the remaining subsets are used for training. The optimal parameters are selected based on the average performance of five times experimental results on the training set. The performance on the testing set with the optimal parameters is utilized to evaluate the performance of the model.
\subsubsection{Experimental results}
\vspace*{5pt}
The fifteen multi-label datasets from UCI machine learning repository\footnote{http://archive.ics.uci.edu/ml/datasets.php} are used as multi-task datasets by treating different labels as different objectives. Their statistics are shown in Table \ref{tablex1}. RBF kernel is employed in these benchmark experiments.
The experimental results of seven algorithms on these benchmark datasets are shown in Table \ref{tablex2}, and the optimal parameters used in experiments are listed in Table \ref{tablex3}. The bold values represent the best accuracy in Table \ref{tablex2}.
\setlength{\tabcolsep}{10mm}
\begin{table*}[!htbp]
	\centering	
	\renewcommand\arraystretch{1.2}
	\caption{The statistics of fifteen benchmark datasets.}\label{tablex1}
	\begin{tabular}{lrrr}
		\hline
		Datasets&$^{\#}$Attributes&$^{\#}$Instances&$^{\#}$Tasks\\
		\hline
		Yeast&104&160&3\\
		Student&5&240&3\\
		Abalone&7&240&3\\
		Corel5k&500&240&3\\
		Scene&295&240&3\\
		Bookmark&2151&120&3\\
		Isolet-ab&242&480&5\\
		Emotion&72&480&6\\
		CAL500&69&120&3\\
		Genbase&1186&120&3\\
		Monk&6&291&3\\
		Flag&19&567&7\\
		Delicious&501&120&3\\
		Mediamill&121&240&3\\
		Recreation&607&240&3\\
		\hline		
	\end{tabular}
\end{table*}

\setlength{\tabcolsep}{2pt}
\begin{table*}[!htbp]	
	\renewcommand\arraystretch{1.2}
	\centering
	\caption{The performance comparison of seven algorithms on fifteen benchmark datasets.}\label{tablex2}
	\resizebox{\textwidth}{!}{
		\begin{tabular}{lllllllll}
			\hline
			Datasets&TWSVM&NPSVM&DMTSVM&MTPSVM&MTLS-TSVM&MTCTSVM&MTNPSVM\\
			~&Accuracy(\%)&Accuracy(\%)&Accuracy(\%)&Accuracy(\%)&Accuracy(\%)&Accuracy(\%)&Accuracy(\%)\\
			~&Time(s)&Time(s)&Time(s)&Time(s)&Time(s)&Time(s)&Time(s)\\
			\hline
			Yeast&66.17&67.83&65.83&68.33&64.83&65.83&\textbf{70.07}\\
			~&0.06&0.19&0.19&0.010&0.17&0.15&0.29\\
			Student&61.67&64.33&62.00&63.00&60.67&64.00&\textbf{64.67}\\
			~&0.05&0.13&0.02&0.001&0.02&0.02&0.07\\
			Abalone&79.33&81.33&80.00&79.33&81.00&80.67&\textbf{81.67}\\
			~&0.06&0.14&0.03&0.010&0.01&0.02&0.08\\
			Corel5k&59.33&64.67&64.00&66.67&65.00&65.67&\textbf{68.00}\\
			~&0.06&0.14&0.03&0.001&0.03&0.04&0.06\\
			Scene&70.33&72.00&93.00&92.33&94.00&94.00&\textbf{94.67}\\
			~&0.06&0.14&0.03&0.001&0.03&0.04&0.06\\
			Bookmark&96.75&96.24&97.75&\textbf{98.75}&98.00&\textbf{98.75}&96.86\\
			~&0.06&0.15&0.02&0.001&0.02&0.06&0.12\\
			Isolet&99.00&99.17&99.30&99.49&\textbf{99.50}&\textbf{99.50}&\textbf{99.50}\\
			~&0.04&0.12&0.01&0.010&0.01&0.25&0.27\\
			Emotion&77.99&\textbf{80.30}&77.15&79.00&76.33&76.91&78.69\\
			~&0.05&0.12&0.03&0.007&0.01&0.19&0.21\\
			CAL500&76.67&80.00&77.33&79.33&77.33&79.33&\textbf{80.00}\\
			~&0.08&0.16&0.01&0.001&0.01&0.01&0.03\\
			Genbase&92.49&94.87&89.83&92.49&92.49&92.49&\textbf{94.90}\\
			~&0.07&0.15&0.02&0.001&0.01&0.04&0.02\\
			Monk&\textbf{86.66}&85.55&85.02&83.72&83.67&84.54&86.37\\
			~&0.04&0.07&0.01&0.001&0.01&0.04&0.07\\
			Flag&73.81&\textbf{75.61}&73.38&74.92&73.72&74.29&75.37\\
			~&0.03&0.12&0.03&0.010&0.01&0.28&0.34\\
			Delicious&61.33&69.33&67.33&68.67&63.33&68.68&\textbf{70.67}\\
			~&0.06&0.14&0.01&0.001&0.01&0.02&0.02\\
			Mediamill&70.83&73.33&73.33&74.17&63.33&67.50&\textbf{75.00}\\
			~&0.07&0.19&0.03&0.001&0.03&0.03&0.04\\
			Recreation&51.00&54.00&51.67&53.33&51.67&54.33&\textbf{54.67}\\
			&0.08&0.18&0.04&0.003&0.05&0.08&0.23\\
			\hline
	\end{tabular}}
\end{table*}
\begin{table*}[!htbp]	
	\centering
	\caption{The optimal parameters of seven algorithms used in the experiments on fifteen benchmark datasets.}\label{tablex3}
	\resizebox{\textwidth}{!}{
		\begin{tabular}{lllllllll}
			\hline
			Datasets&TSVM&NPSVM&DMTSVM&MTPSVM&MTLS-TSVM&MTCTSVM&MTNPSVM\\
			~&$(c,\rho)$&$(c_{1},c_{2},\delta,\epsilon)$&$(c,\rho,\delta)$&$(c,\rho,\delta)$&$(c,\rho,\delta)$&$(c,g,\rho,\delta)$&$(\rho,c_{1},c_{2},\delta,\epsilon)$\\
			\hline
			Yeast&$(2^{-3},2^{3})$&$(2^{-1},2^{-1},2^{3},0.1)$&$(2^{-3},2^{2},2^{3})$&$(1,2^{-1},2^{3})$&$(2^{-3},1,2^{3})$&$(2^{-3},2^{-2},2^{-3},2^{3})$&$(2^{-1},1,2^{-2},2^{3},0.1)$\\
			Student&$(1,2^{3})$&$(2^{2},2^{2},2^{3},0.1)$&$(1,2^{2},2^{3})$&$(2^{2},2^{1},2^{3})$&$(2^{-1},2^{2},2^{3})$&$(1,2^{2},2^{2},2^{3})$&$(2^{-3},2^{-1},1,2^{2},0.1)$\\
			Abalone&$(2^{2},2^{3})$&$(2^{3},1,2^{2},0.1)$&$(2^{-3},2^{3},2^{2})$&$(2^{2},2^{1},2^{-1})$&$(2^{1},2^{3},2^{1})$&$(2^{-1},2^{2},2^{-3},2^{1})$&$(2^{3},2^{2},2^{-1},2^{1},0.1)$\\
			Corel5k&$(2^{-3},2^{3})$&$(2^{-3},2^{-3},2^{2},0.1)$&$(2^{-3},2^{3},2^{3})$&$(2^{-2},2^{1},2^{3})$&$(2^{-3},2^{3},2^{3})$&$(2^{-3},2^{1},2^{3},2^{3})$&$(2^{-3},2^{-3},2^{-3},2^{3},0.1)$\\
			Scene&$(2^{-3},2^{3})$&$(2^{-1},2^{-1},2^{3},0.1)$&$(2^{-3},2^{3},2^{3})$&$(2^{-1},2^{3},2^{3})$&$(2^{-1},2^{3},2^{3})$&$(2^{-3},2^{1},2^{2},2^{3})$&$(2^{-3},2^{-3},2^{-3},2^{2},0.1)$\\
			Bookmark&$(2^{-3},2^{-3})$&$(2^{-3},2^{-3},2^{2},0.1)$&$(2^{-3},1,2^{-3})$&$(2^{-3},2^{1},2^{-3})$&$(2^{-3},2^{1},2^{-3})$&$(2^{-3},2^{-3},2^{-1},1)$&$(2^{-2},2^{-3},2^{-3},2^{3},0.1)$\\
			Isolet-ab&$(2^{-3},2^{3})$&$(2^{-3},2^{-2},2^{2},0.1)$&$(2^{-3},1,2^{1})$&$(2^{-2},1,2^{1})$&$(2^{},2^{},2^{})$&$(2^{-3},2^{2},2^{3},2^{1})$&$(2^{-3},2^{-3},2^{-3},2^{1},0.1)$\\
			Emotion&$(2^{-3},1)$&$(1,2^{-2},1,0.1)$&$(2^{-3},2^{-2},1)$&$(2^{-3},2^{-3},1)$&$(2^{-3},2^{-3},1)$&$(2^{-3},1,2^{-3},1)$&$(2^{-3},2^{-3},2^{-3},2^{1},0.1)$\\
			CAL500&$(2^{-3},2^{-3})$&$(2^{-2},2^{-2},2^{2},0.1)$&$(2^{-3},2^{-1},2^{2})$&$(2^{-3},2^{-3},2^{3})$&$(2^{-3},2^{1},2^{2})$&$(2^{-3},2^{1},2^{-3},2^{2})$&$(1,2^{-3},2^{-2},2^{3},0.1)$\\
			Genbase&$(2^{-3},2^{3})$&$(2^{1},2^{1},2^{3},0.1)$&$(2^{-3},2^{-3},2^{3})$&$(2^{-2},2^{-3},2^{3})$&$(2^{-3},2^{-3},2^{3})$&$(2^{-3},2^{3},2^{1},2^{3})$&$(2^{1},1,2^{-1},2^{3},0.1)$\\
			Monk&$(2^{1},2^{2})$&$(2^{3},2^{3},1,0.1)$&$(2^{-3},2^{-1},2^{2})$&$(2^{3},2^{-3},2^{1})$&$(2^{3},2^{-3},2^{1})$&$(2^{-3},2^{3},2^{-2},2^{2})$&$(2^{1},2^{1},2^{1},2^{1},0.1)$\\
			Flag&$(2^{-2},2^{2})$&$(2^{1},2^{-1},1,0.1)$&$(2^{-1},2^{2},2^{3})$&$(2^{-1},2^{2},2^{2})$&$(1,2^{-3},2^{3})$&$(2^{-2},2^{3},1,2^{2})$&$(2^{-1},1,2^{-2},2^{1},0.1)$\\
			Delicious&$(2^{-3},2^{1})$&$(2^{-2},1,2^{3},0.1)$&$(2^{-3},2^{1},2^{3})$&$(2^{1},1,2^{3})$&$(2^{-3},2^{3},2^{2})$&$(2^{-3},2^{-1},2^{-1},2^{3})$&$(2^{3},2^{-3},2^{-1},2^{3},0.1)$\\
			Mediamill&$(2^{-3},2^{1})$&$(2^{-1},2^{1},2^{2},0.1)$&$(2^{-3},2^{3},2^{1})$&$(2^{2},2^{3},2^{1})$&$(2^{-3},2^{2},2^{1})$&$(2^{-3},2^{-1},2^{2},2^{1})$&$(2^{-3},2^{-3},2^{-3},2^{1},0.1)$\\
			Recreation&$(2^{-1},2^{3})$&$(2^{-2},1,2^{3},0.1)$&$(2^{-3},2^{-1},2^{2})$&$(2^{3},2^{-1},2^{3})$&$(1,1,2^{-2})$&$(1,2^{1},2^{-1},2^{3})$&$(2^{-3},2^{-3},2^{-2},2^{3},0.1)$\\
			\hline
	\end{tabular}}
\end{table*}

In terms of accuracy, MTNPSVM performs better than the remaining methods on two thirds of the datasets. Compared to the STL methods, although MTNPSVM has a lower computational efficiency due to the necessary to train multiple tasks simultaneously, it also achieves better generalization performance as a result. Compared to the other MTL methods, MTNPSVM performs the best on most of the benchmark datasets. This also indicates that the $\epsilon$-insensitive loss function not only has higher theoretical sparsity than the square loss function, but is also more conducive to the construction of the decision hyperplane. In terms of the running time, MTNPSVM takes longer time since it needs to handle larger scale problems than DMTSVM and MTCTSVM. The better computational efficiency of MTLS-TSVM and MTPSVM is due to the fact that they only need to deal with linear programming problems, but it is worth noting that there is no sparsity in these two models.
\subsubsection{Friedman test}
\vspace*{5pt}
It is not intuitively observable here that MTNPSVM performs better than the other models in Table \ref{tablex2}. To differentiate the performance of the seven algorithms, the Friedman test is introduced as a non-parametric post-hoc test. The average ranks of the seven algorithms with respect to accuracy are tabulated in Table \ref{tablex4}.
\setlength{\tabcolsep}{2pt}
\begin{table*}[!htbp]	
	\renewcommand\arraystretch{1.2}
	\centering
	\caption{Average rank on classification accuracy of seven algorithms.}\label{tablex4}
	\resizebox{\textwidth}{!}{
		\begin{tabular}{lllllllll}
			\hline
			Datasets&TWSVM&NPSVM&DMTSVM&MTPSVM&MTLS-TSVM&MTCTSVM&MTNPSVM\\
			\hline
			Yeast&4&3&6&2&7&5&1\\
			Student&6&2&5&4&7&3&1\\
			Abalone&6.5&2&5&6.5&3&4&1\\
			Corel5k&7&5&6&2&4&3&1\\
			Scene&7&6&4&5&2.5&2.5&1\\
			Bookmark&6&7&4&1.5&3&1.5&5\\
			Isolet-ab&7&6&4&5&2&2&2\\
			Emotion&4&1&5&2&7&6&3\\
			CAL500&7&1.5&5.5&3.5&5.5&3.5&1.5\\
			Genbase&6&2&7&4&4&4&1\\
			Monk&1&2&4&6&7&5&2\\
			Flag&5&1&7&3&6&4&2\\
			Delicious&7&2&5&4&6&3&1\\
			Mediamill&5&3.5&3.5&2&7&6&1\\
			Recreation&7&3&5.5&4&5.5&2&1\\
			\hline
			Average rank&5.70&3.30&5.10&3.63&5.10&3.63&1.63\\
			\hline
	\end{tabular}}
\end{table*}

Under the null hypothesis, all algorithms are equivalent. The Friedman statistic \cite{fried} can be computed as follows:
\begin{eqnarray}
	\label{friedman}
	\chi_{F}^{2}=\frac{12N}{k(k+1)}\left[\sum_{j}R_{j}^{2}-\frac{k(k+1)^{2}}{4}\right],
\end{eqnarray}
where the $k$ and $N$ represent the number of algorithms and datasets, respectively, and the $R_{j}$ denotes the average rank of the $j$-th algorithm on all datasets. Since the original Friedman statistic above was too conservative, the new statistic is derived as follows:
\begin{eqnarray}
\label{fdistri}
F_{F}=\frac{(N-1)\chi_{F}^{2}}{N(k-1)-\chi_{F}^{2}},
\end{eqnarray}
where the $F_{F}$ obeys to the $F$-distribution with $k-1$ and $(k-1)(N-1)$ degrees of freedom. The $\chi_{F}^{2}=39.8915$ and $F_{F}=11.1454$ can be obtained according to (\ref{friedman}) and (\ref{fdistri}). Here the $F_{F}$ obeys to the $F$-distribution with $(6,84)$. When the level of significance $\alpha$=$0.05$ the critical value of $F(6,84)$ is $2.20$, and similarly $2.56$ at $\alpha$=$0.025$. The $F_{F}$ is much larger than the critical value which means that there are very significant differences between the seven algorithms. It should be noted that the average rank of MTNPSVM is much lower than the remaining algorithms, which proves that MTNPSVM outperforms the remaining methods.
\subsection{Analysis of model}\label{analysis}
In this subsection, the model is further analyzed. Firstly, two solution methods are compared to demonstrate the efficiency of ADMM algorithm used in above solving process. Then performance influence of task size, property of parameter $\epsilon$, convergence of algorithm, and parameter sensitivity are further analyzed. The grid-search strategy and 5-fold cross-validation are performed in this subsection.
\subsubsection{Solution method}\label{sec_con}
\vspace*{5pt}
``quadprog" function in MATLAB is often leveraged to settle the quadratic programming problems. To demonstrate the validity of the ADMM algorithm, the performance of MTNPSVM solved by ADMM algorithm and ``quadprog" function in MATLAB are shown in the Table \ref{table3}. Here three datasets $landmine$\footnote{http://people.ee.duke.edu/lcarin/LandmineData.zip}, $Letter$\footnote{http://archive.ics.uci.edu/ml/datasets.php}, and $Spambase$\footnote{http://www.ecmlpkdd2006.org/challenge.html} are often used to evaluate multi-task learning. The specific information can also be found in \cite{roughv}. As shown, it can be found that the ADMM algorithm can speed up the training speed while only a slight change in the training accuracy. Although the computational time is still more than other models, the computational efficiency has been significantly improved compared to the previous ``quadprog" function.
\setlength{\tabcolsep}{4pt}
\begin{table*}[!htbp]
	\renewcommand\arraystretch{1.2}	
	\centering
	\caption{Performance comparison of different solving methods on three benchmark datasets.}\label{table3}
	\begin{tabular}{cccccccccccc}\hline
		&Method&&Landmine&&Letter&&Spambase&&\\
		\cline{3-10}&~&&Accuracy(\%) &Time(s)& Accuracy(\%)
		&Time(s) &Accuracy(\%) &Time(s)\\
		\noalign{\smallskip} \hline \noalign{\smallskip}
		&MTNPSVM-QUAD&&79.60&0.10&96.80&0.21&93.01&0.29\\
		&MTNPSVM-ADMM&&79.80&0.05&96.83&0.12&92.20&0.18\\
		\hline
	\end{tabular}
\end{table*}
\subsubsection{Performance influence of task size}
\begin{figure}[!htbp]	
	\centering
	\subfigure{\includegraphics[width=0.48\textwidth]{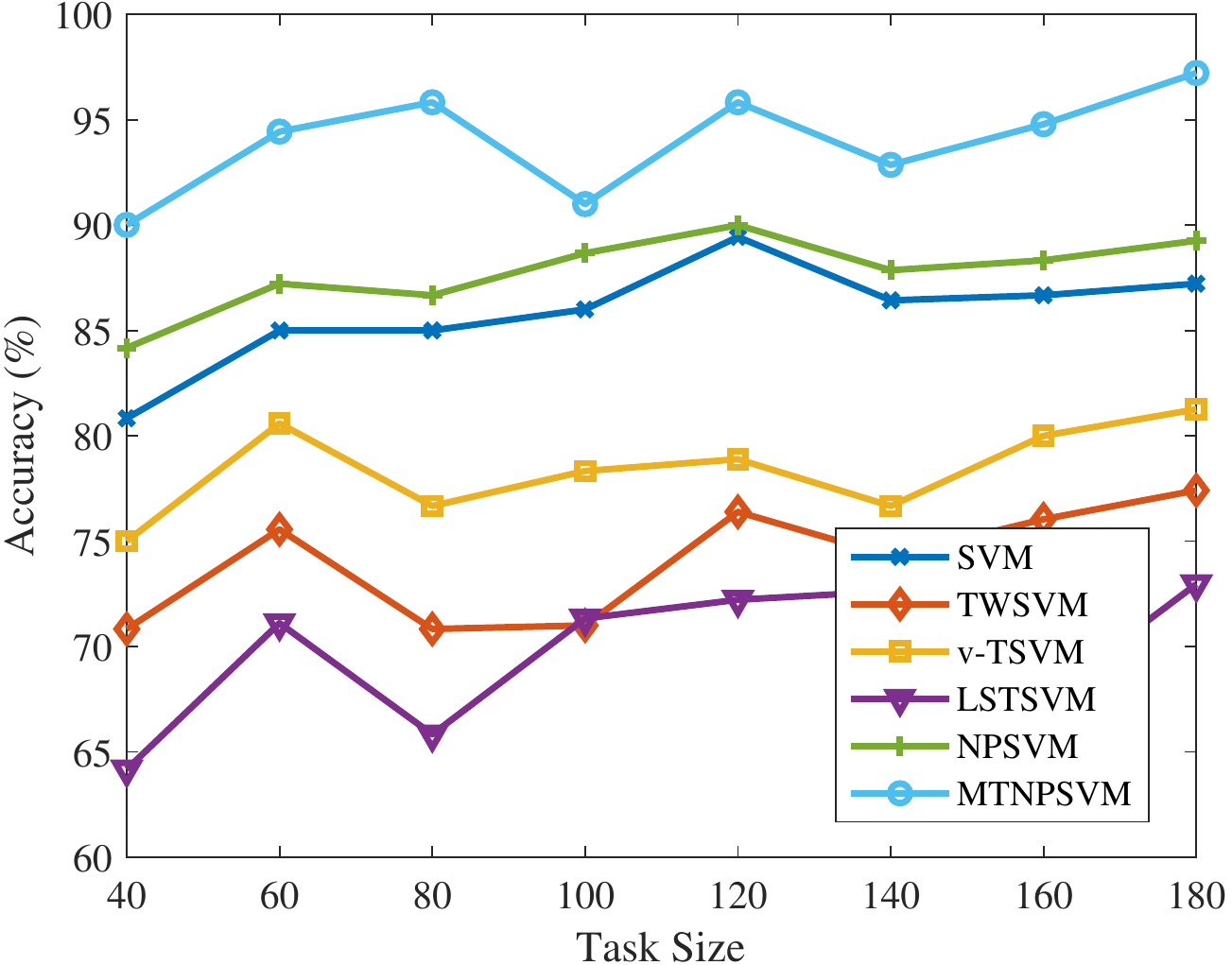}}
	\subfigure{\includegraphics[width=0.48\textwidth]{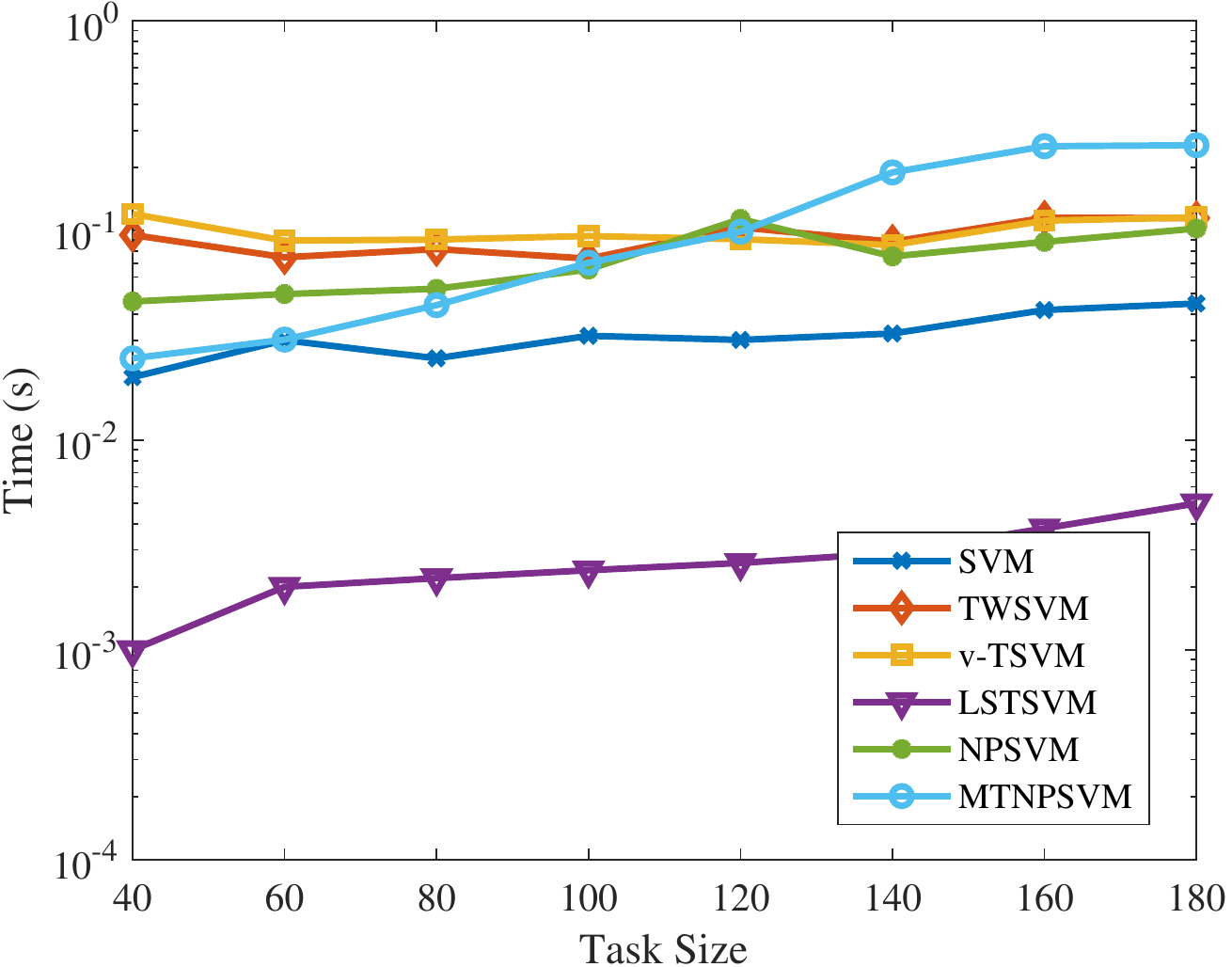}}	
	\caption{The performance comparison between STL methods and MTNPSVM on $spambase$ dataset with RBF kernel.}	
	\label{fig2}
\end{figure}

\begin{figure}[!htbp]	
	\centering
	\subfigure{\includegraphics[width=0.48\textwidth]{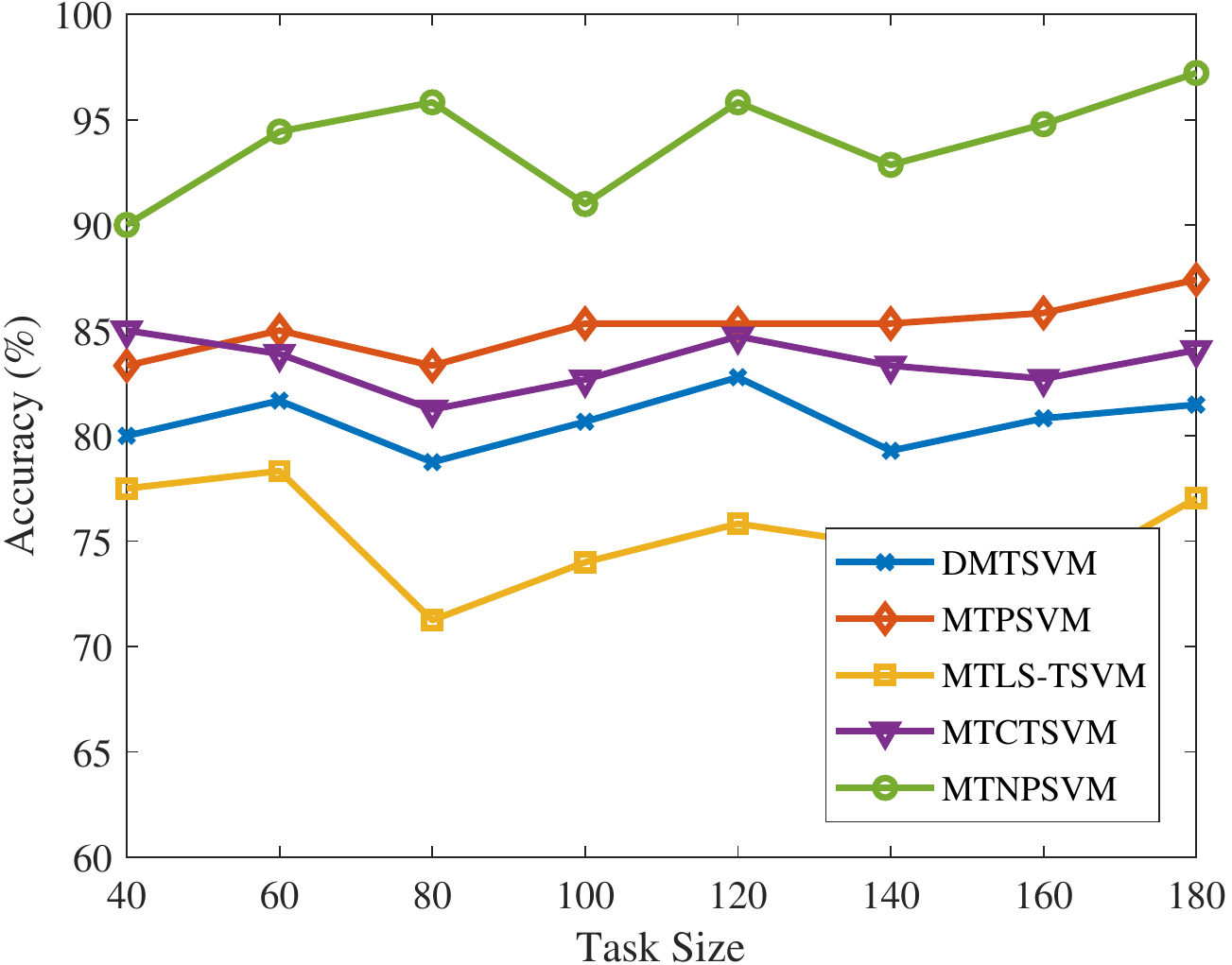}}
	\subfigure{\includegraphics[width=0.48\textwidth]{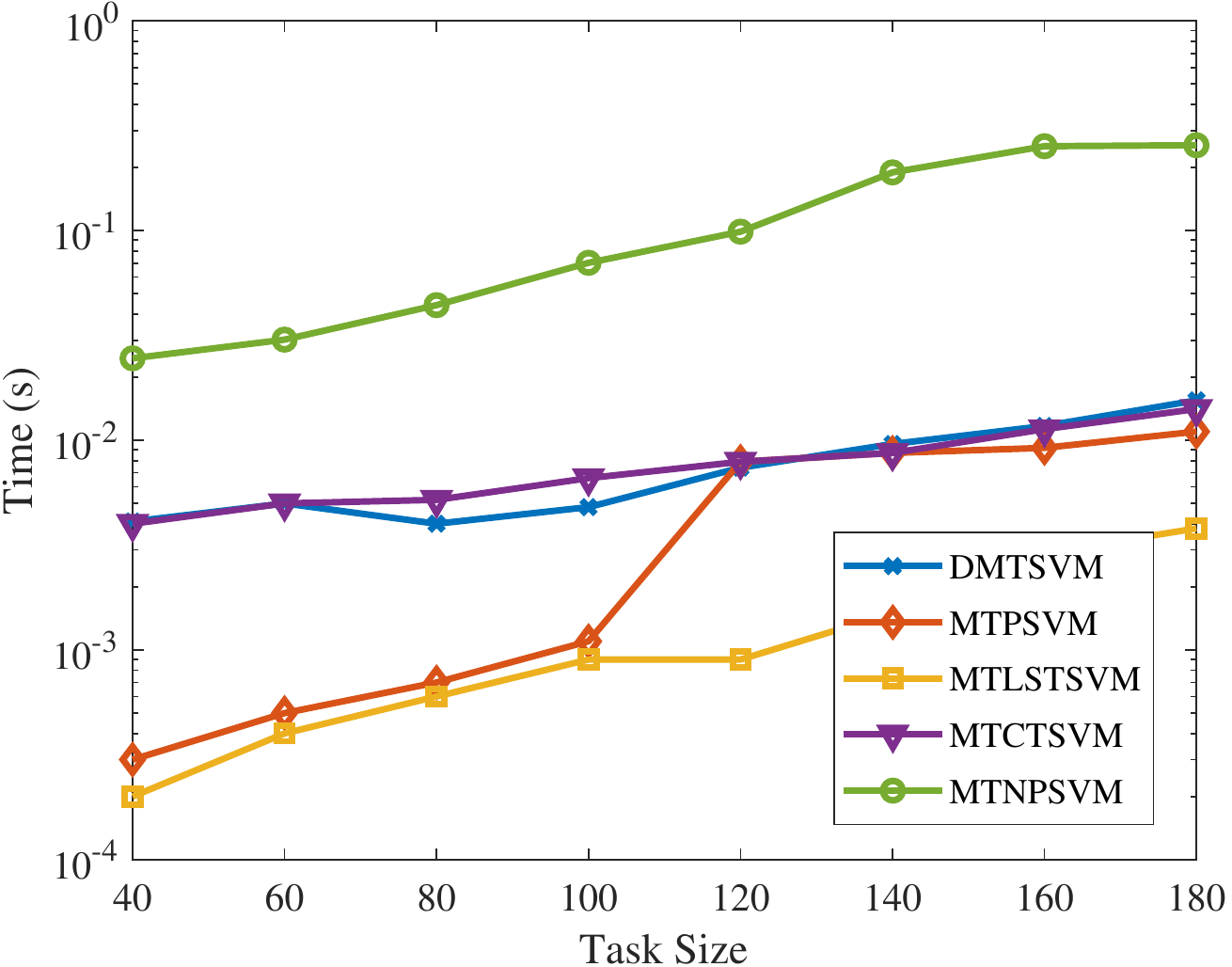}}	
	\caption{The performance comparison between MTL methods and MTNPSVM on $spambase$ dataset with RBF kernel.}	
	\label{fig3}
\end{figure}

\vspace*{5pt}
$Spambase$ dataset is a binary dataset for spam identification, which includes three tasks and each task contains 200 mails and the final data contains 36 features reduced through PCA. In order to further explore the influence of task size on generalization performance, the $Spambase$ dataset is resized to different scales, ranging from 40 to 180. In addition, MTNPSVM is compared with all STL methods and MTL methods, respectively. The experimental results at different scales of task with RBF kernel are displayed in Figs. \ref{fig2} and \ref{fig3}.

In Fig. \ref{fig2}, the experimental results indicate that MTNPSVM has much better performance than other STL methods with the increasing task size. Also it can be found that the prediction accuracy increases roughly with the task size, which indicated that the larger size of task is helpful for us to better discover the intrinsic properties of data. In addition, it can be found that the training duration of all methods rise with the task size, it can be explained that the extended number of samples increases the matrix dimension in programming, thereby aggravating the burden of calculation. In Fig. \ref{fig3}, MTNPSVM has better generalization performance than other MTL methods for different task sizes. Moreover, a similar conclusion to Fig. \ref{fig2} can be drawn, i.e., as the task size gets larger, the testing accuracy gets higher and the computational time gets longer.

By comparing the accuracy of STL methods and MTL methods globally in Figs. \ref{fig2} and \ref{fig3}, the multi-task learning method has more stable and better generalization performance than the STL methods when the sample size is very small, but with the increasing of the number of samples, the gap between the two kinds of methods will become smaller and smaller. It can be explained as follows, single-task learning cannot fully explore the potential information of the samples when the sample size is small, while MTL methods can effectively improve the overall generalization performance by exploring the similar structural information among multiple related tasks. This results in a more obvious advantage of MTL methods. However, as the sample size increases, STL methods can explore the data information using sufficient samples, so the gap between the two types of methods is reduced. Therefore, multi-task learning can fully demonstrate its advantages with small samples. The similar conclusions can be drawn by referring to \cite{vtwin}.
\subsubsection{Property of parameter $\epsilon$}
\vspace*{5pt}
In order to demonstrate the property of parameter $\epsilon$, this subsection carries out experiments on the MTNPSVM with different kernels. Although $\epsilon$ increases the burden of parameter selection, it adds convenience to adjust the sparsity of the dual solution. It can influence the number of support vectors (SVs) by adjusting the value of $\epsilon$. After cross-validation and grid search, the other parameters of the model are fixed as optimal.

\begin{figure}[!htbp]
	\centering
	\subfigure[]{\includegraphics[width=0.45\textwidth]{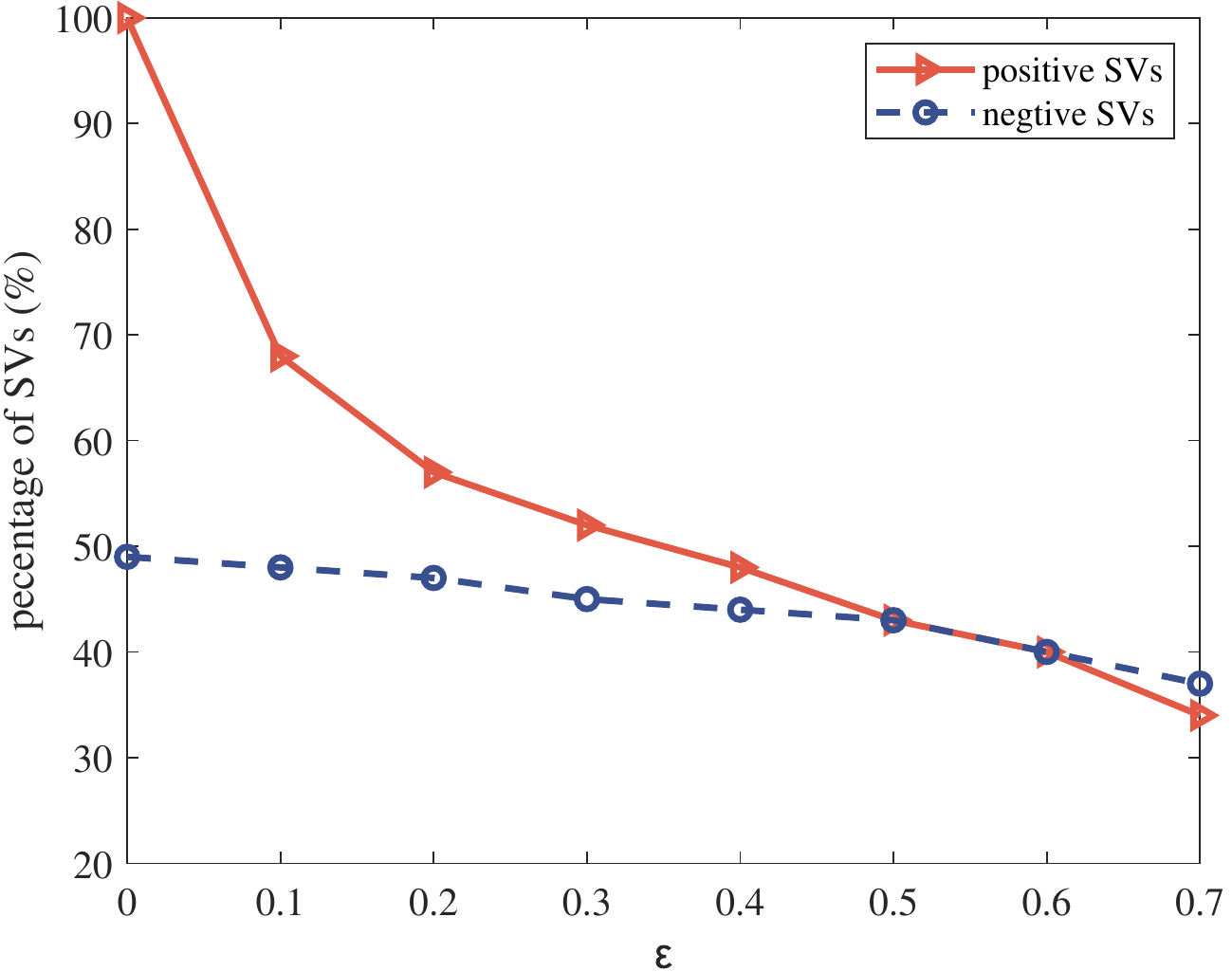}}
	\subfigure[]{\includegraphics[width=0.45\textwidth]{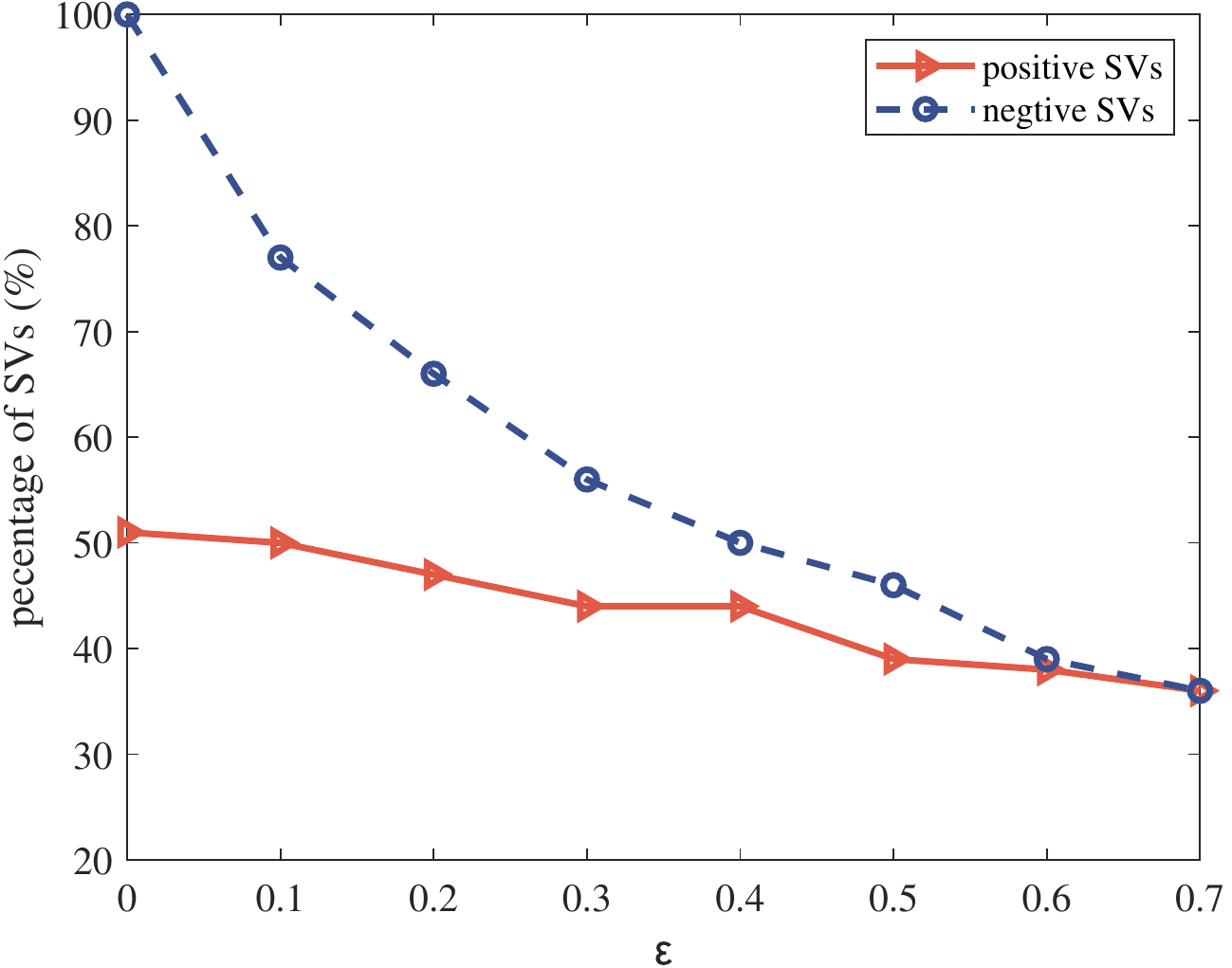}}
	\caption{Sparseness changes in $Spambase$ datasets with the increasing $\epsilon$.}
	\label{figsv1}
\end{figure}

\begin{figure}[!htbp]
	\centering
	\subfigure[]{\includegraphics[width=0.45\textwidth]{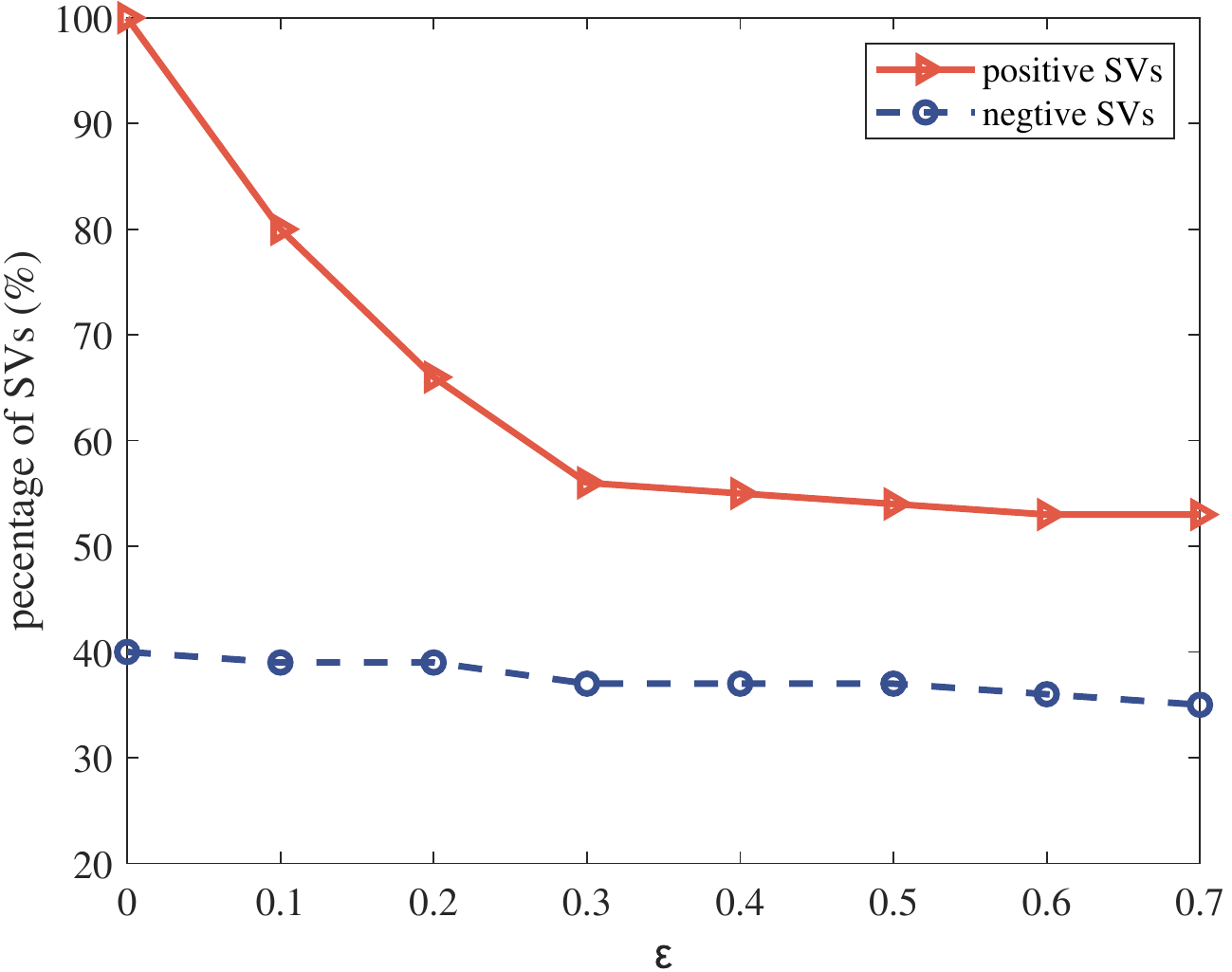}}
	\subfigure[]{\includegraphics[width=0.45\textwidth]{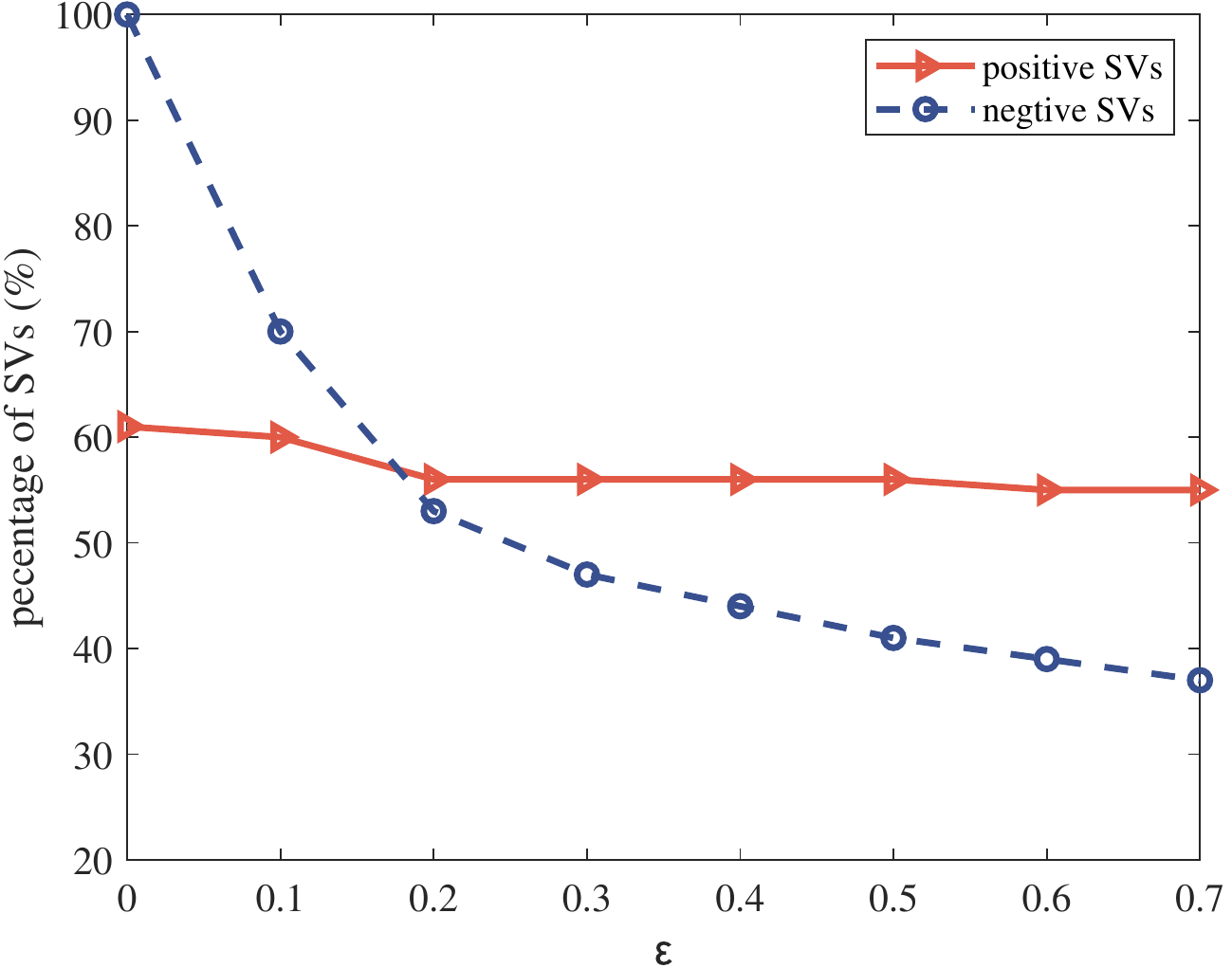}}
	\caption{Sparseness changes in $Landmine$ datasets with the increasing $\epsilon$.}
	\label{figsv2}
\end{figure}

Figs. \ref{figsv1}$\sim$\ref{figsv2} (a), (b) illustrate the variations of SVs in two different QPPs, respectively. In Fig. \ref{figsv1}, while $\epsilon$ goes bigger and the other relevant parameters are remained unchanged, the number of SVs in class itself decreases obviously and less in the other class, so that sparseness increases. Furthermore, the number of SVs in class itself changes greatly which indicates that more positive samples are close to the decision hyperplane. The similar phenomenon on $Landmine$ dataset can be found in Fig. \ref{figsv2}.
\subsubsection{Convergence analysis}
\begin{figure}[!htbp]
	\centering
	\subfigure[$Spambase$]{\includegraphics[width=0.49\textwidth]{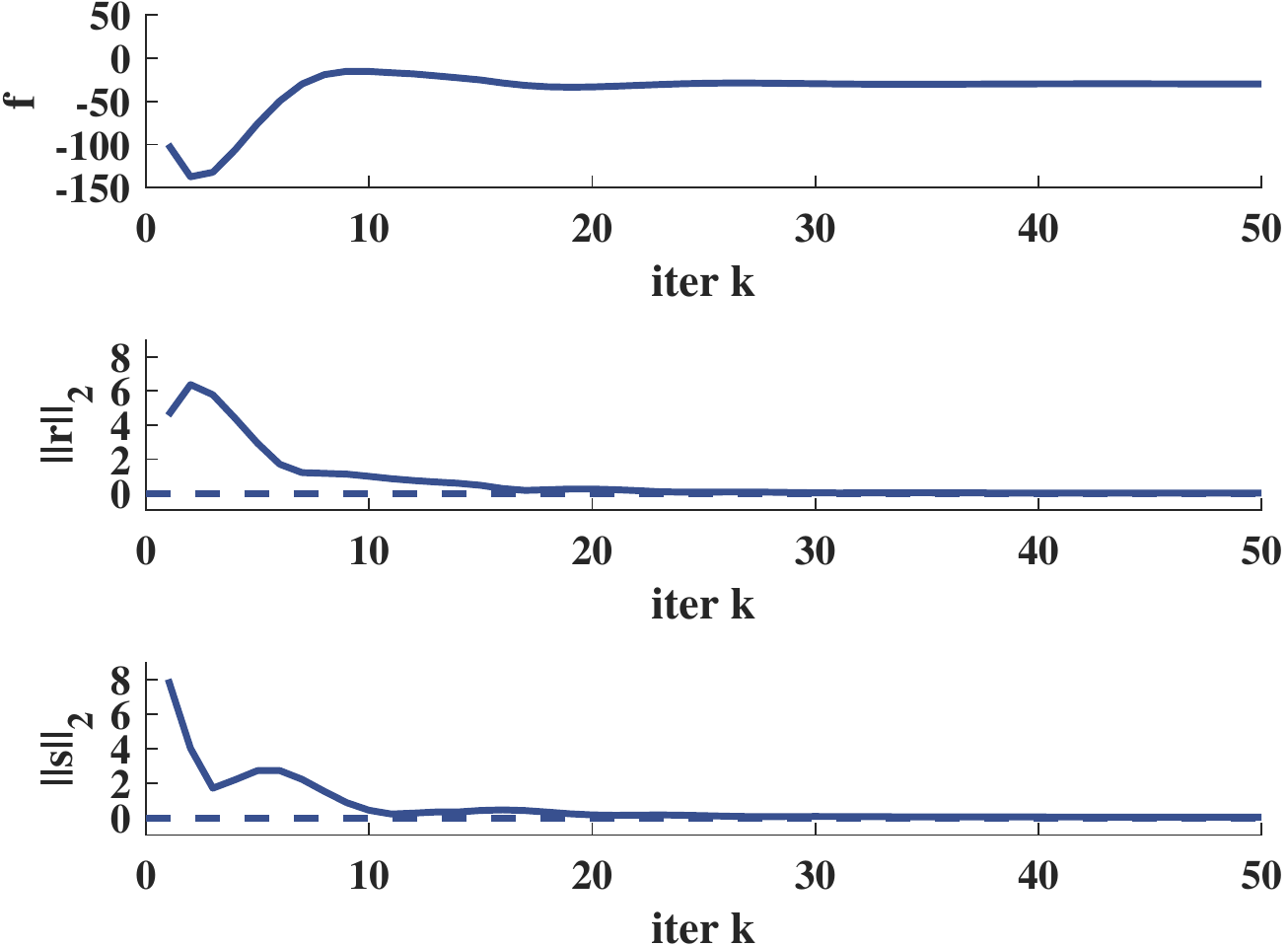}}
	\subfigure[$Spambase$]{\includegraphics[width=0.49\textwidth]{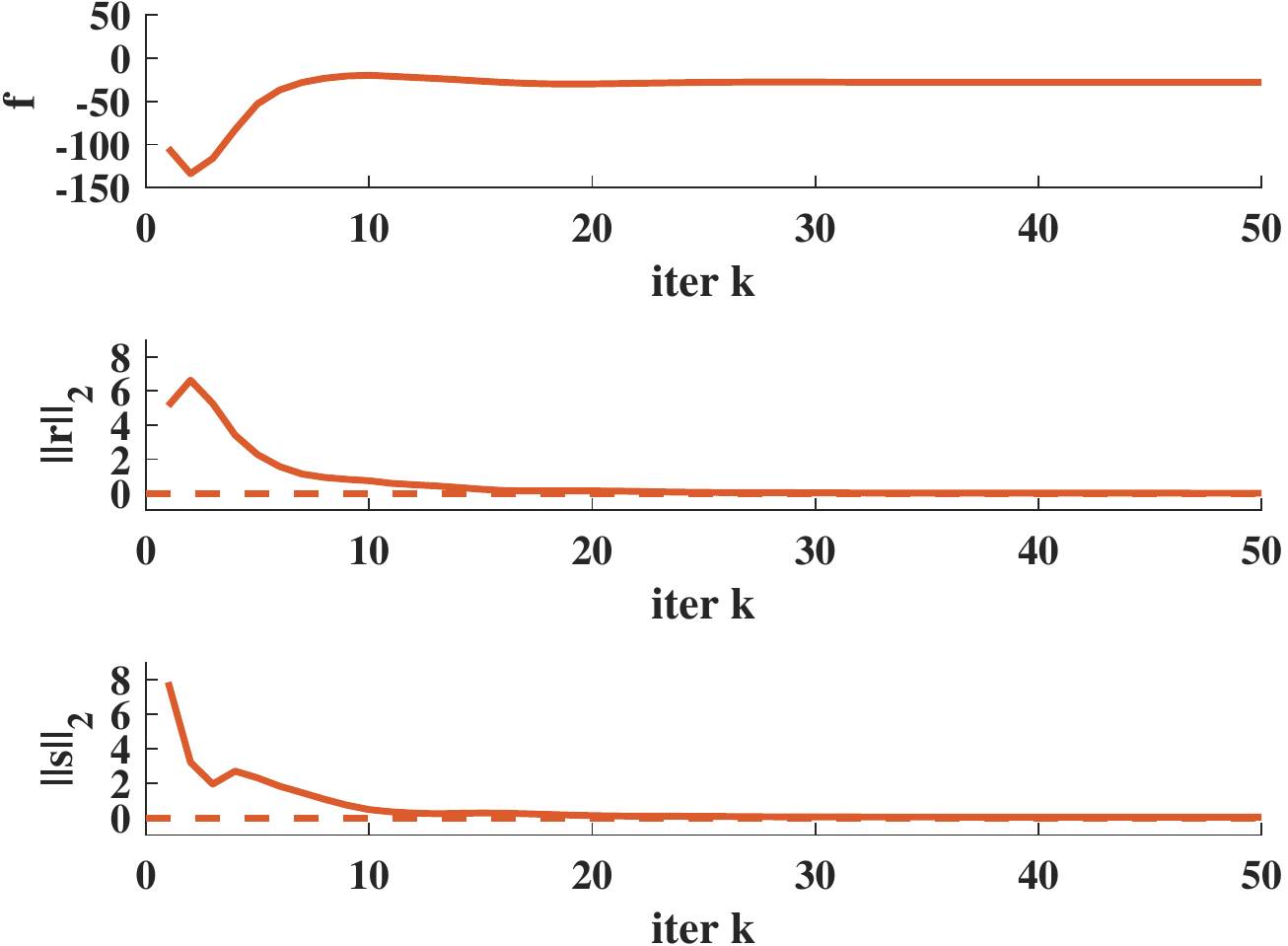}}
	\subfigure[$Landmine$]{\includegraphics[width=0.49\textwidth]{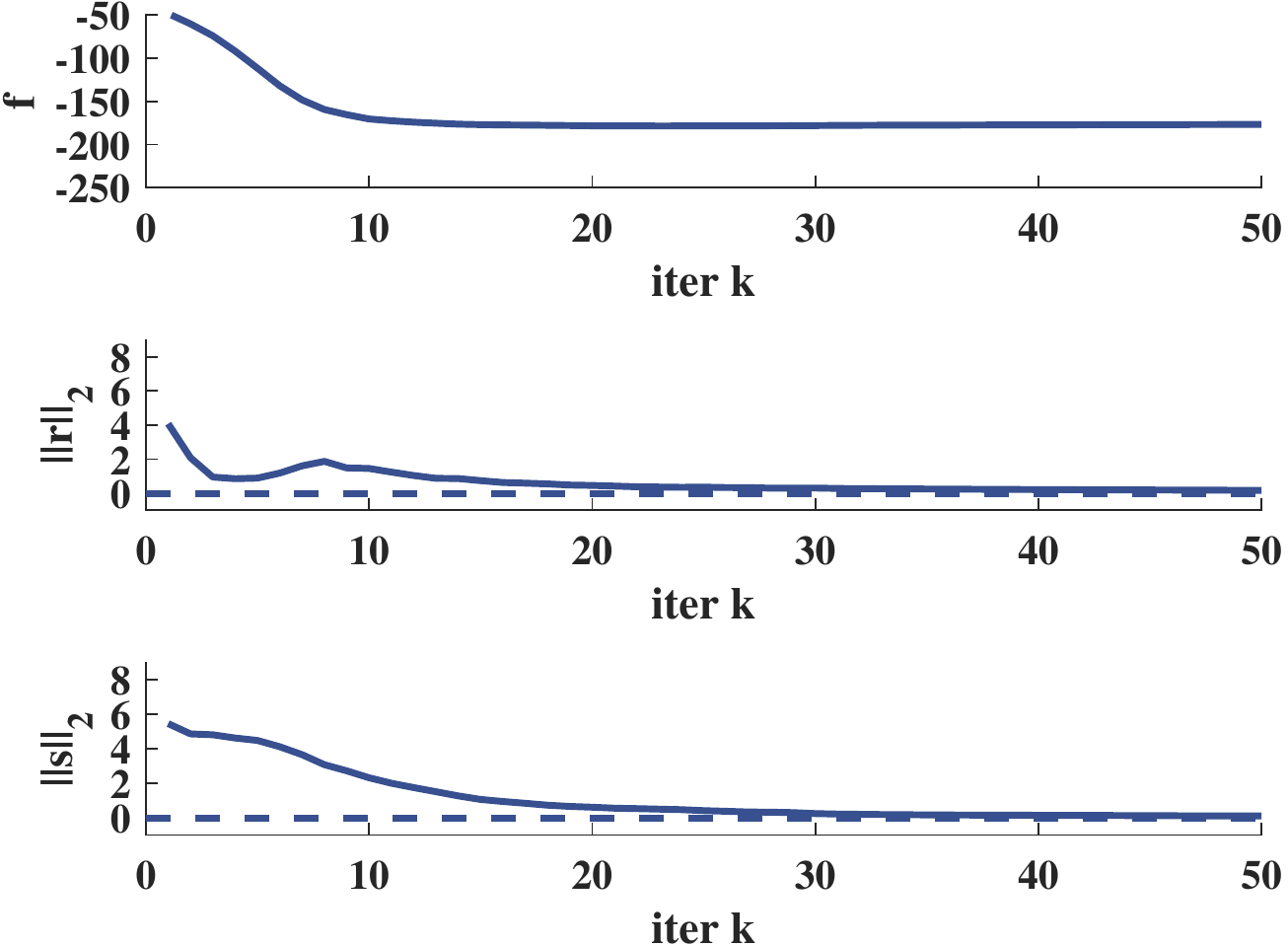}}
	\subfigure[$Landmine$]{\includegraphics[width=0.49\textwidth]{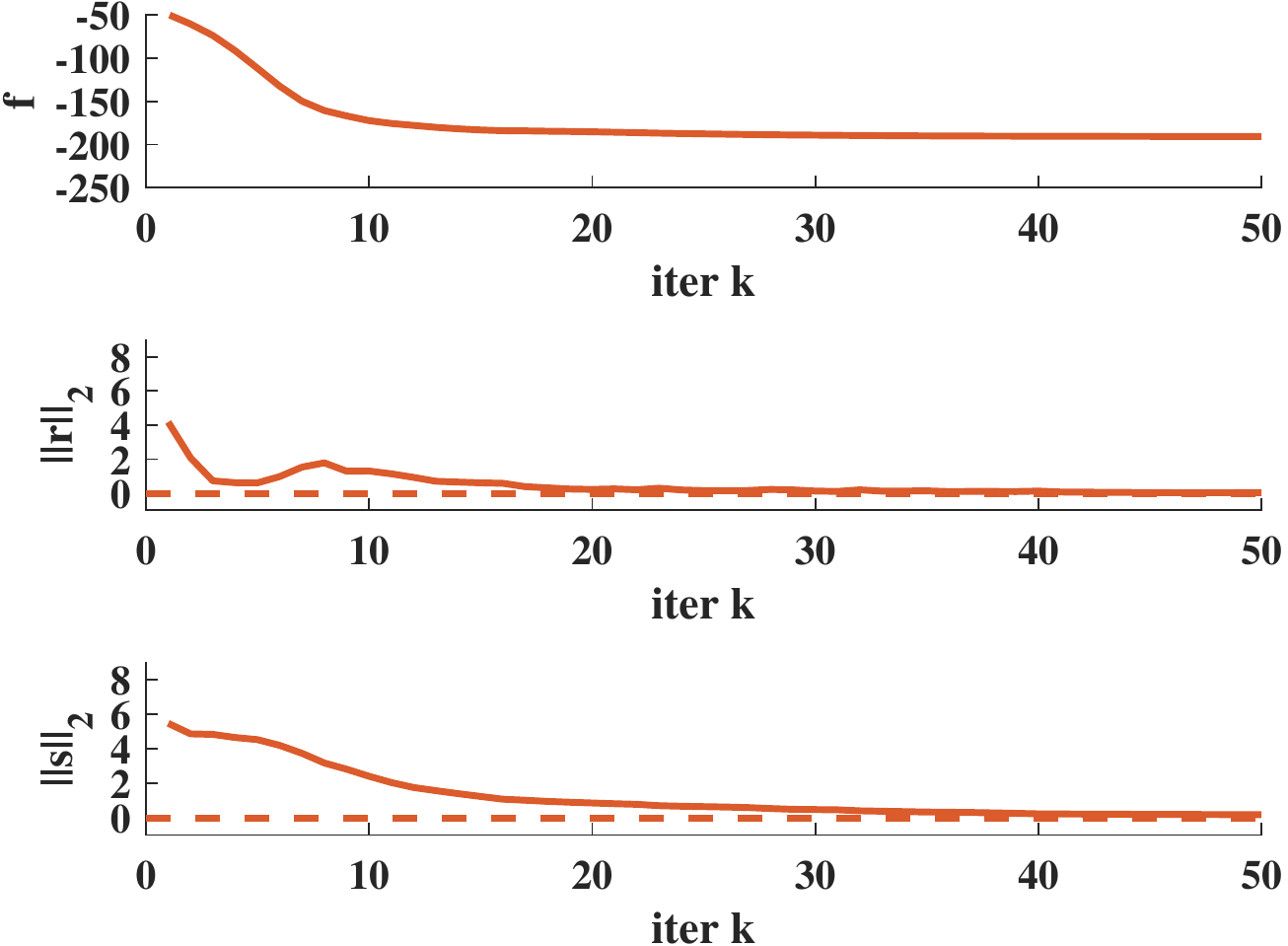}}
	\caption{Convergence of ADMM in two datasets with RBF kernel. (a), (c) and (b), (d) represent two different programming problems in the proposed model, as (\ref{2009}), (\ref{2010}). Objective function value, primal residual, and dual residual are represented by the solid lines, respectively. The convergence of the approximation values are represented by the dashed lines.}
	\label{figiter}
\end{figure}

\vspace*{5pt}
To better understand the convergence process of the ADMM, the objective function $f$, primal residual $\|r\|_{2}$, and dual residual $\|s\|_{2}$ as several crucial indicators, their variation curves are displayed in Fig. \ref{figiter} with RBF kernel. The hyperparameters are fixed as the optimal parameters obtained by 5-fold cross-validation and grid search.

 As the number of iterations increases, it can be found that primal residual $\|r\|_{2}$ and dual residual $\|s\|_{2}$ will be close to 0 and vary slightly, while the objective function values $f$ in problems (\ref{1018}) and (\ref{1019}) tend to a fixed value after the certain iterations. The experimental results reveal that MTNPSVM can be solved well by ADMM and finally converges efficiently.
\subsubsection{Parameter sensitivity}
\begin{figure}[!htbp]
	\centering
	\subfigure[$\delta$=0.125]{\includegraphics[width=0.3\textwidth]{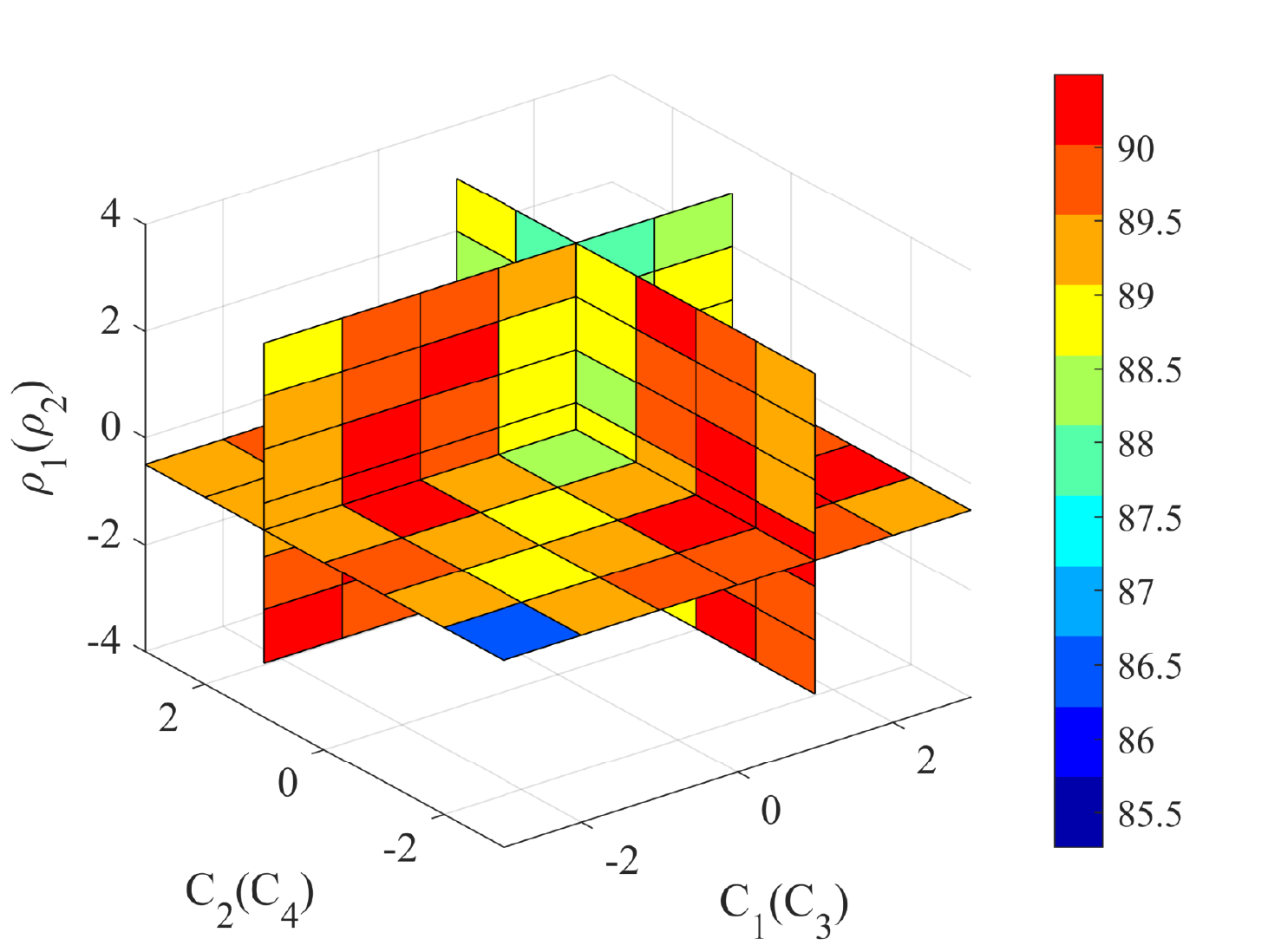}}
	\subfigure[$\delta$=1]{\includegraphics[width=0.3\textwidth]{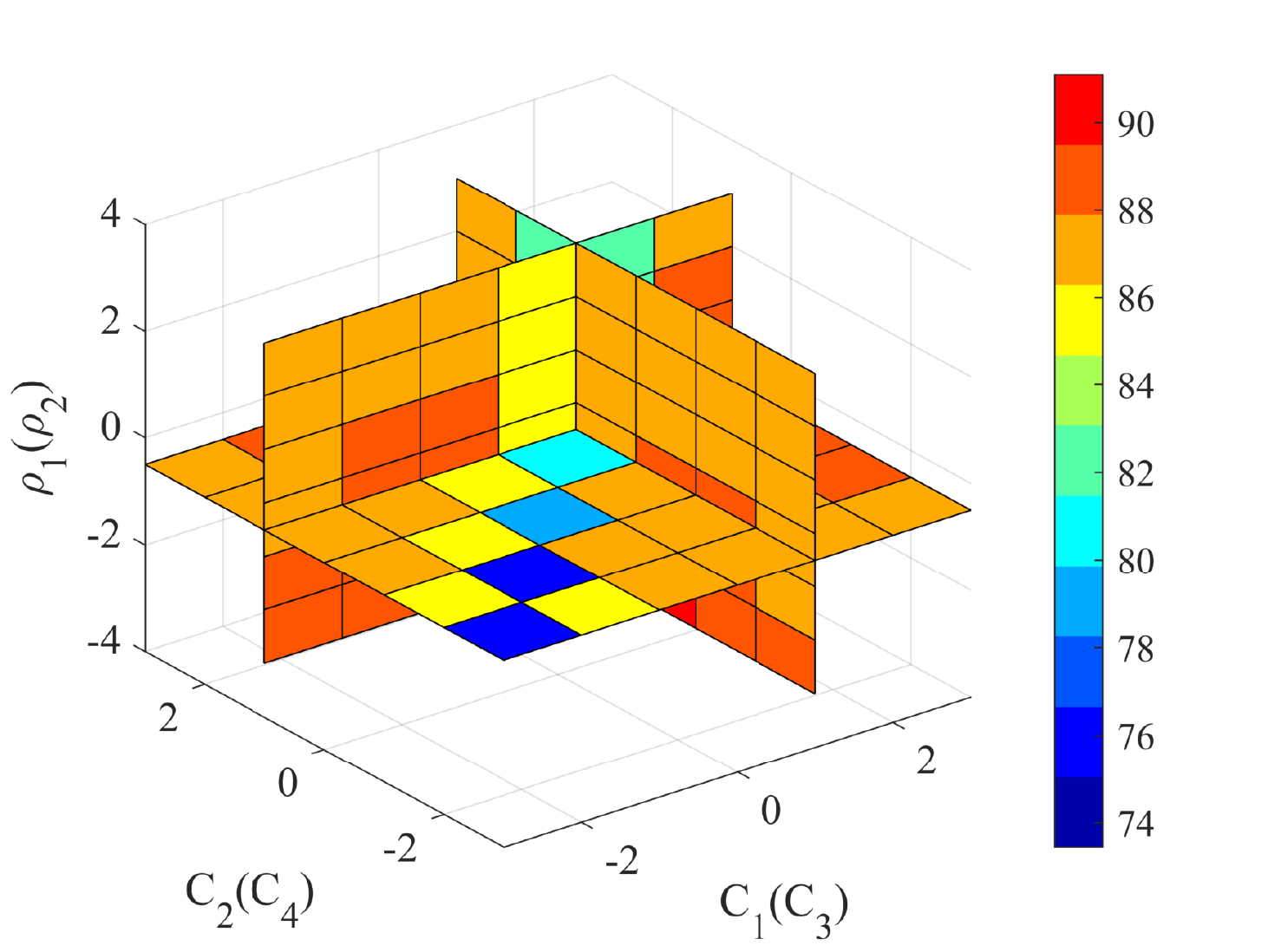}}
	\subfigure[$\delta$=8]{\includegraphics[width=0.3\textwidth]{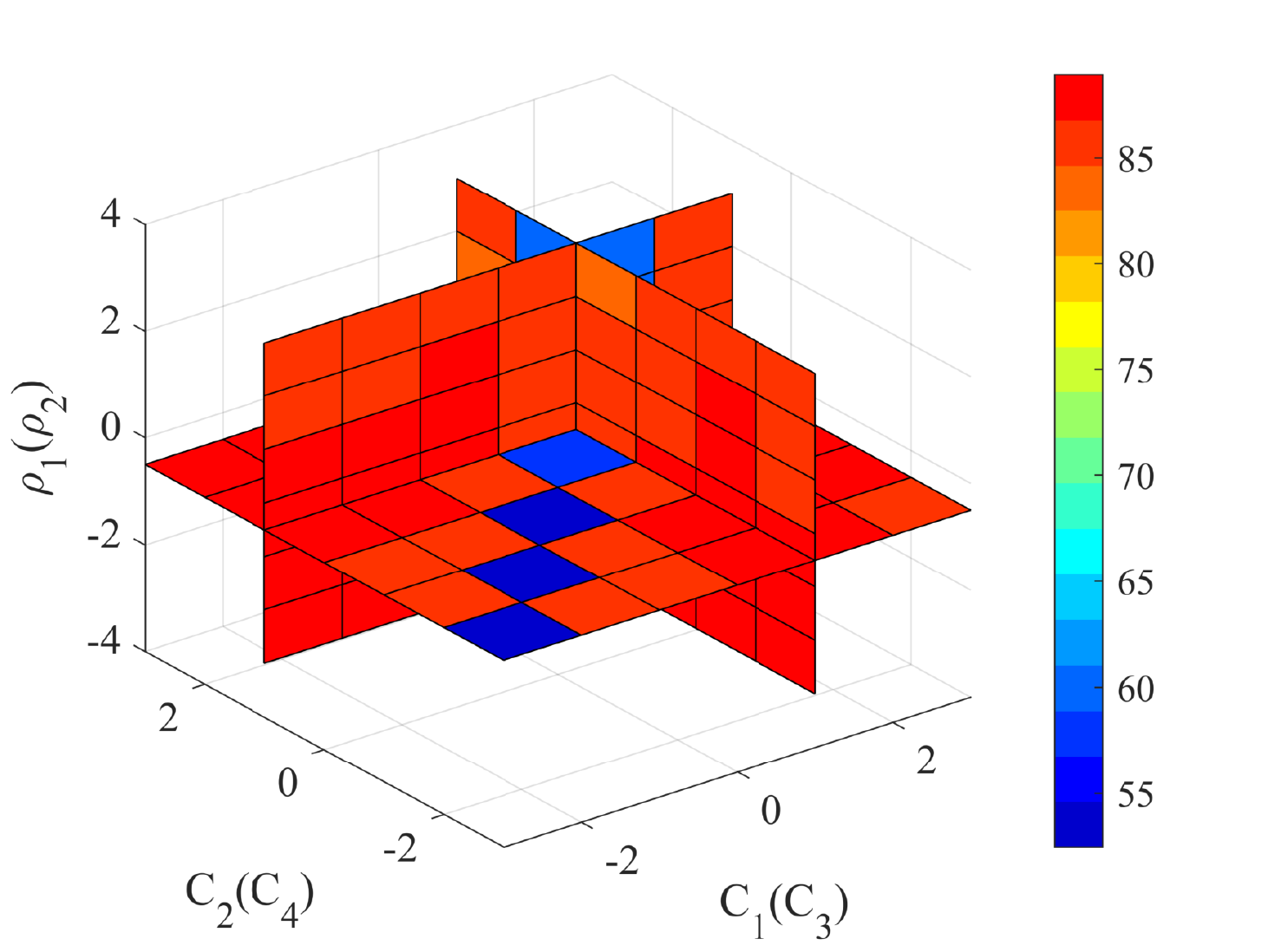}}
	\caption{Parameter sensitivity on $Spambase$ with different kernel parameters.}
	\label{para_1}
\end{figure}

\begin{figure}[!htbp]
	\centering
	\subfigure[$\delta$=0.125]{\includegraphics[width=0.3\textwidth]{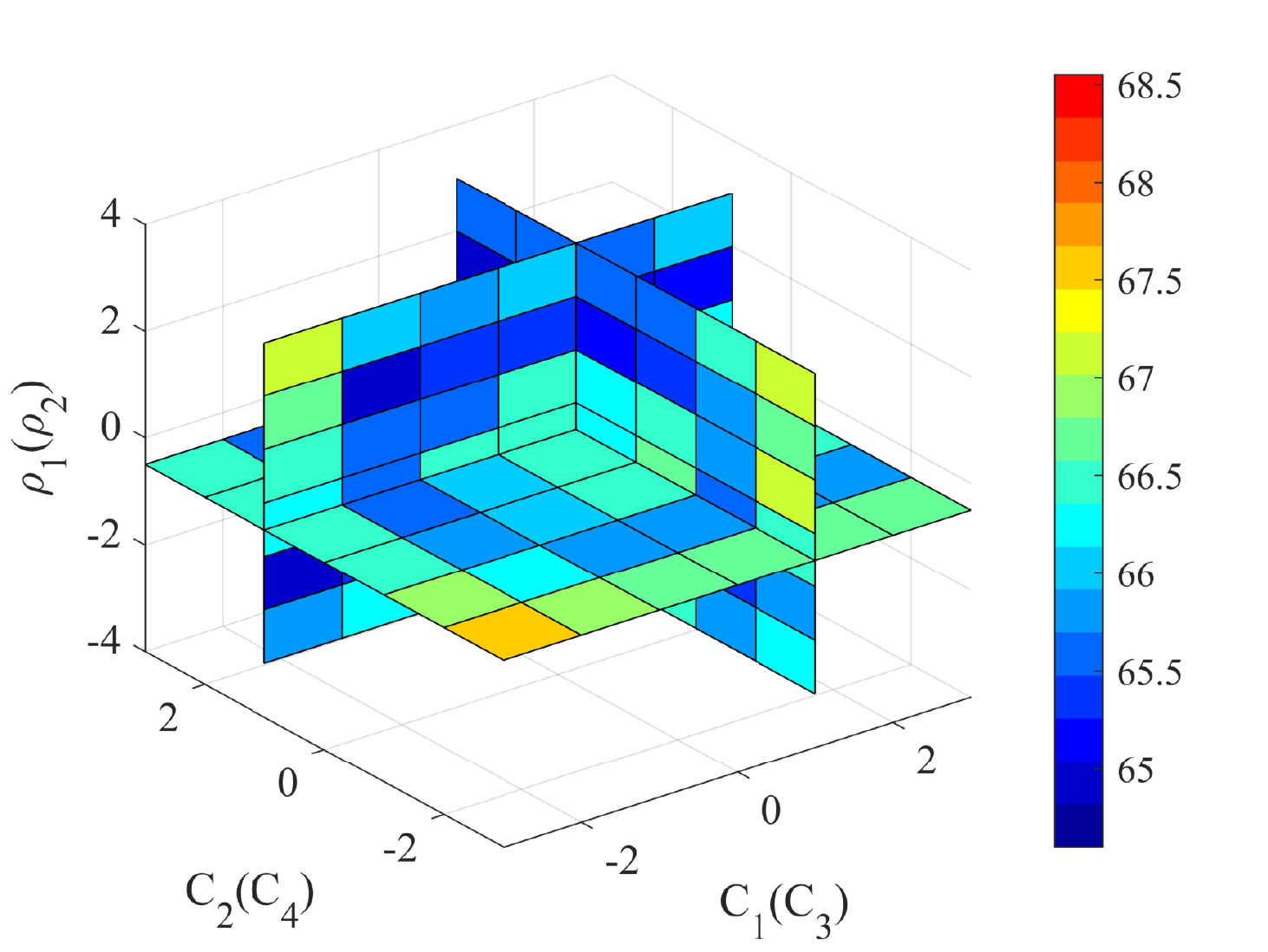}}
	\subfigure[$\delta$=1]{\includegraphics[width=0.3\textwidth]{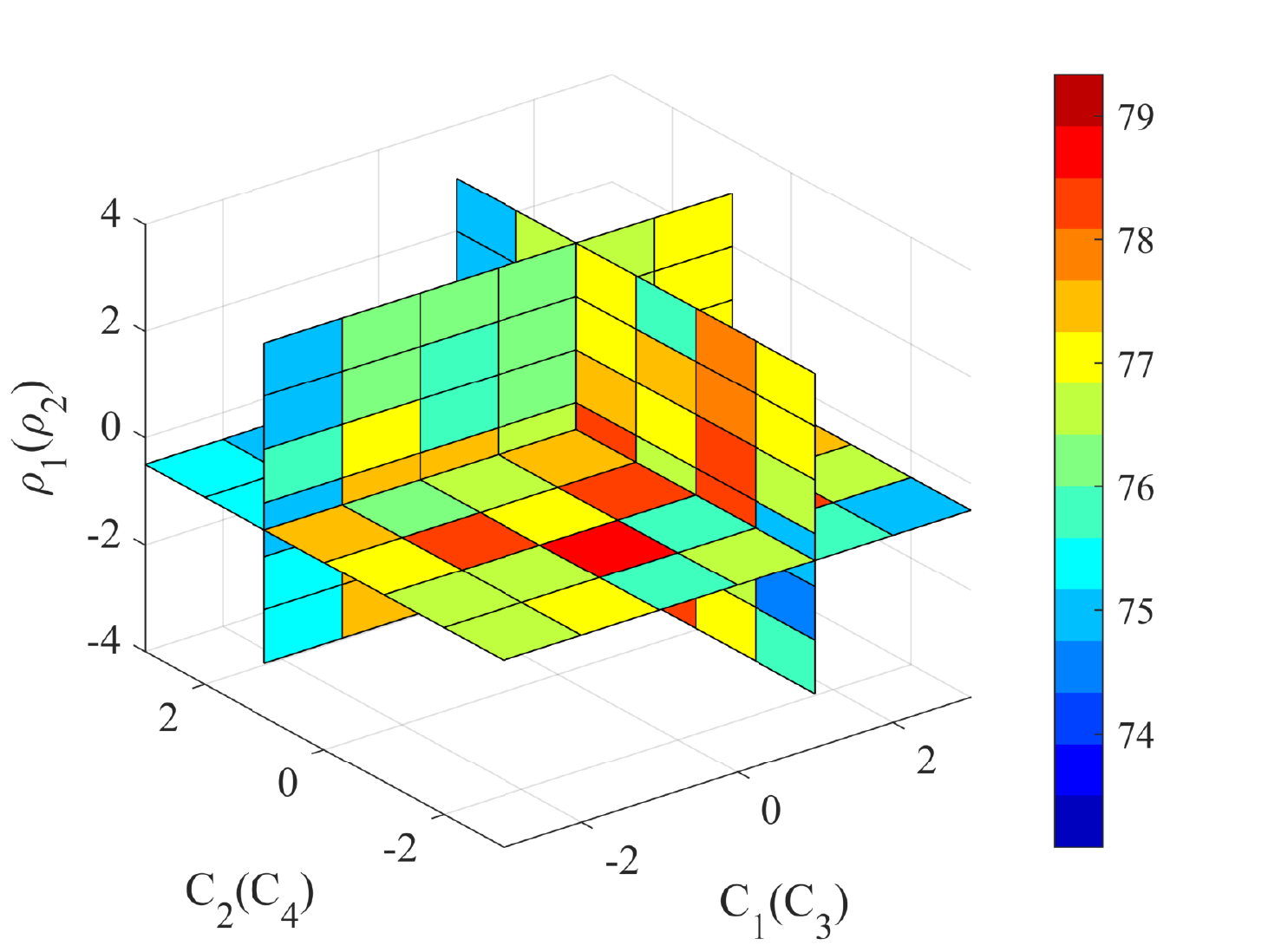}}
	\subfigure[$\delta$=8]{\includegraphics[width=0.3\textwidth]{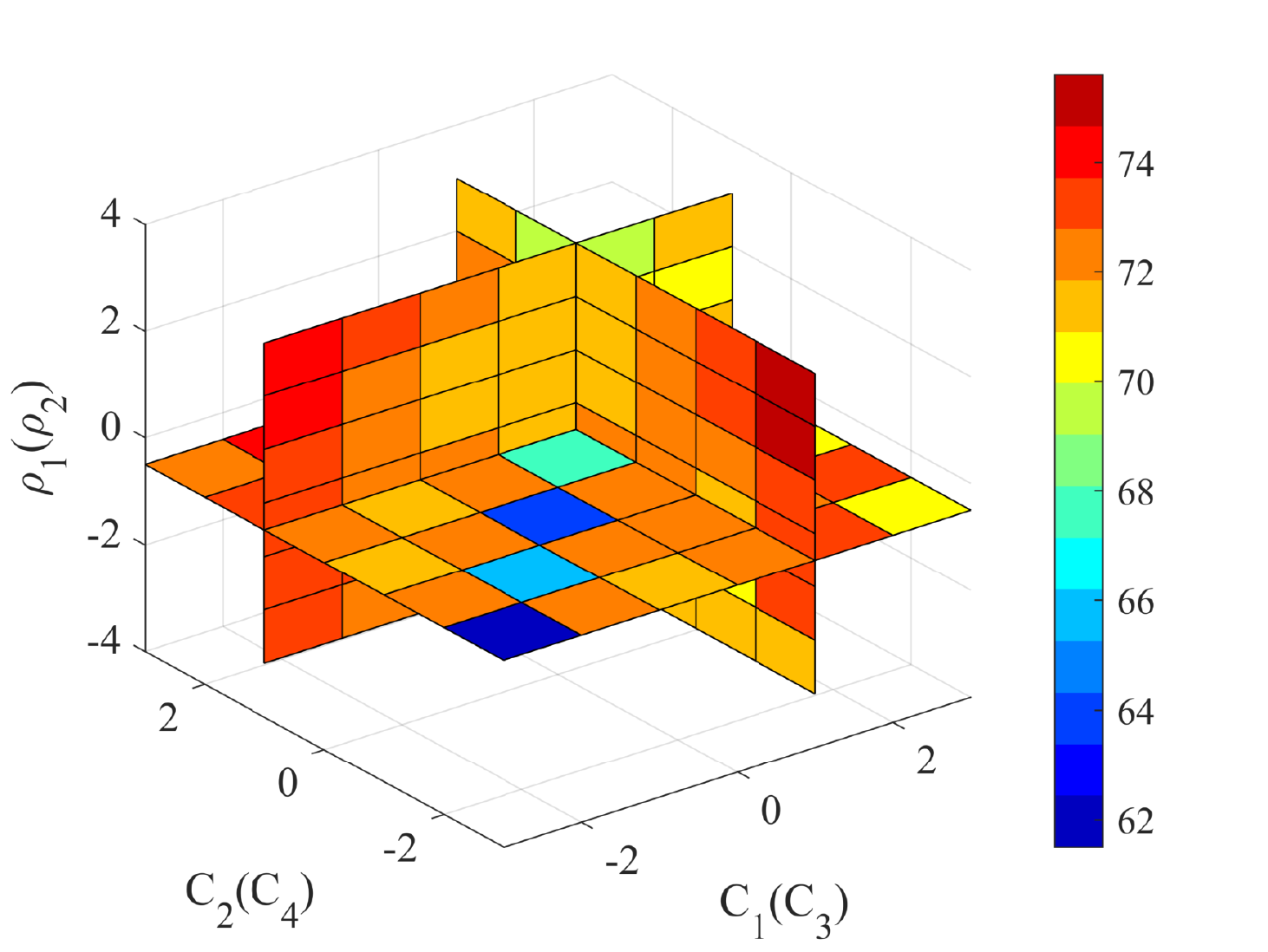}}
	\caption{Parameter sensitivity on $Landmine$ with different kernel parameters.}
	\label{para_2}
\end{figure}

\vspace*{5pt}
In order to further explore the effect of the main parameters on the final generalization performance, the parameters $\rho_{1}$($\rho_{2}$), $C_{1}$($C_{3}$) and $C_{2}$($C_{4}$) are chosen to conduct the numerical experiments on two benchmark datasets with the rest of parameter fixed, the scale of color indicates the accuracy, and the three axes represent three different parameters. The same grid search and cross-validation as in the previous experiments are also executed. In order to investigate the effect of the model sensitivity to three types of different parameters, the RBF kernel function with different kernel parameter values is applied in the Figs. \ref{para_1} and \ref{para_2}, respectively. The experimental results are analyzed to reach the following conclusions: (a) the model is becoming increasingly more insensitive to the $\rho$ with the increasing $\delta$. (b) MTNPSVM has comparable sensitivity to parameter $C_{1}(C_{3})$ and parameter $C_{2}(C_{4})$.

\subsection{Experiments on image datasets}
\begin{figure}[!htbp]	
	\centering	
	\includegraphics[width=0.8\textwidth,height=0.3\textheight]{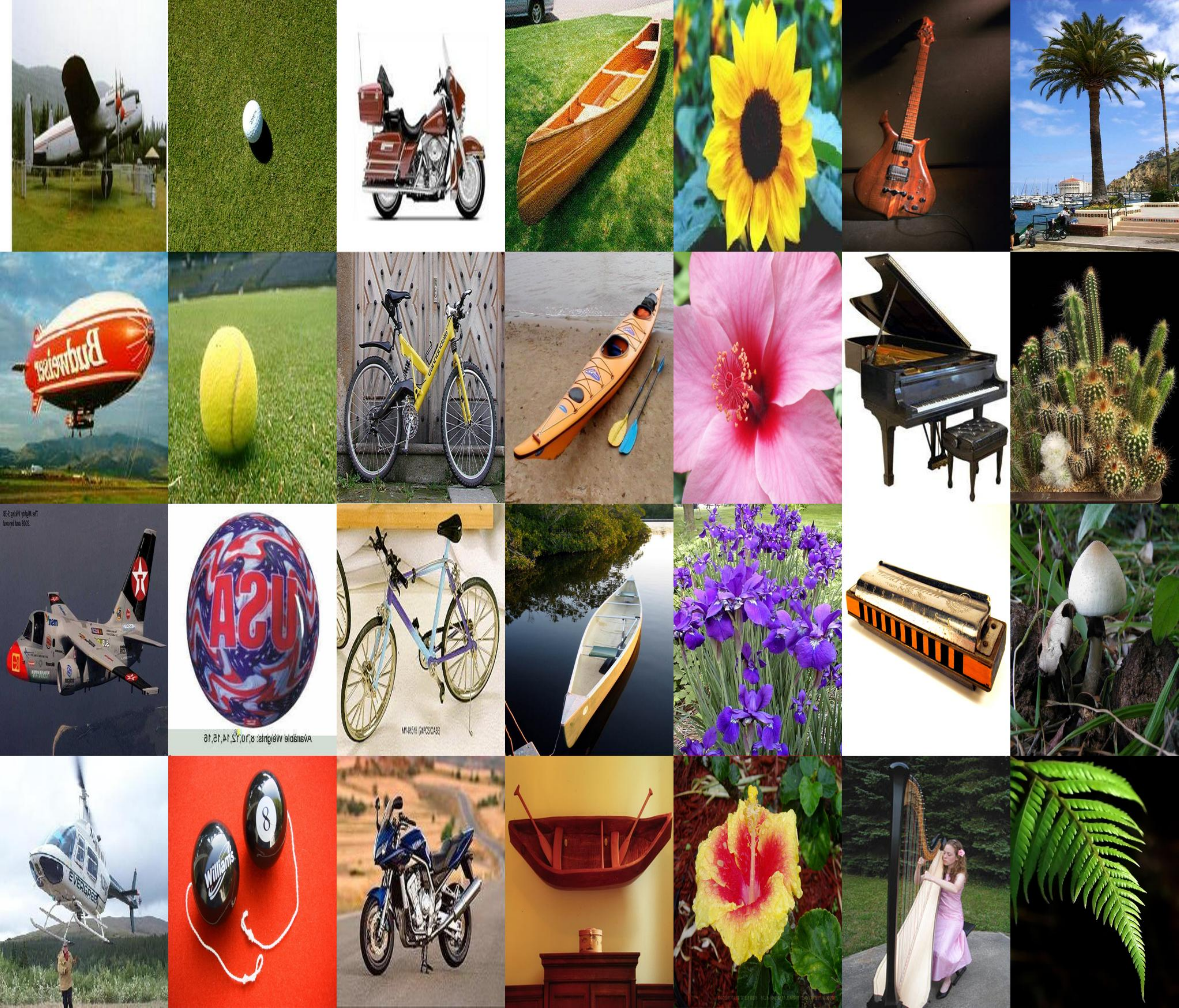}
	\caption{Caltech image dataset.}
	\label{fig4}
\end{figure}
To verify the performance of the MTNPSVM in comparison with the other MTL methods, this subsection searches for two very well-known Caltech image repositories including Caltech 101 and the Caltech 256 \cite{cal101,cal256}. Caltech 101 includes 102 classes of image and a background class, each of which has about 50 images. Caltech 256 has 256 classes of image and a background class, each of which has no less than 80 images. The samples in background class are not in any of image categories, so it can be viewed as negative class. To transform Caltech images into multiple datasets with similar structural information, based on the architecture information, the related subclasses are synthesized into a large superclass. Some categories of pictures are displayed in Fig. \ref{fig4}, each superclass contains from 3 to 6 subclasses. It can be found that each column of pictures has a similar feature information. For instance, in the first column, their aerocrafts all contain the cabin, wings, tail, etc., so they can be seen as a superclass. Eventually each subclass is mixed with negative samples. identiting samples belonging to similar superclasses in different subclasses can be viewed as a set of related tasks. In Caltech 101, five multi task datasets are synthesized in final, the number of samples in each superclass is selected 50 images, so the final number of each task is 100. Similarly, seven multi-task datasets are combined from Caltech 256. Finally, multi-task learning improves the generalization performance by exploiting the similar structure information between tasks.

\begin{figure}[!htbp]	
	\centering
	\subfigure{\includegraphics[width=0.45\textwidth]{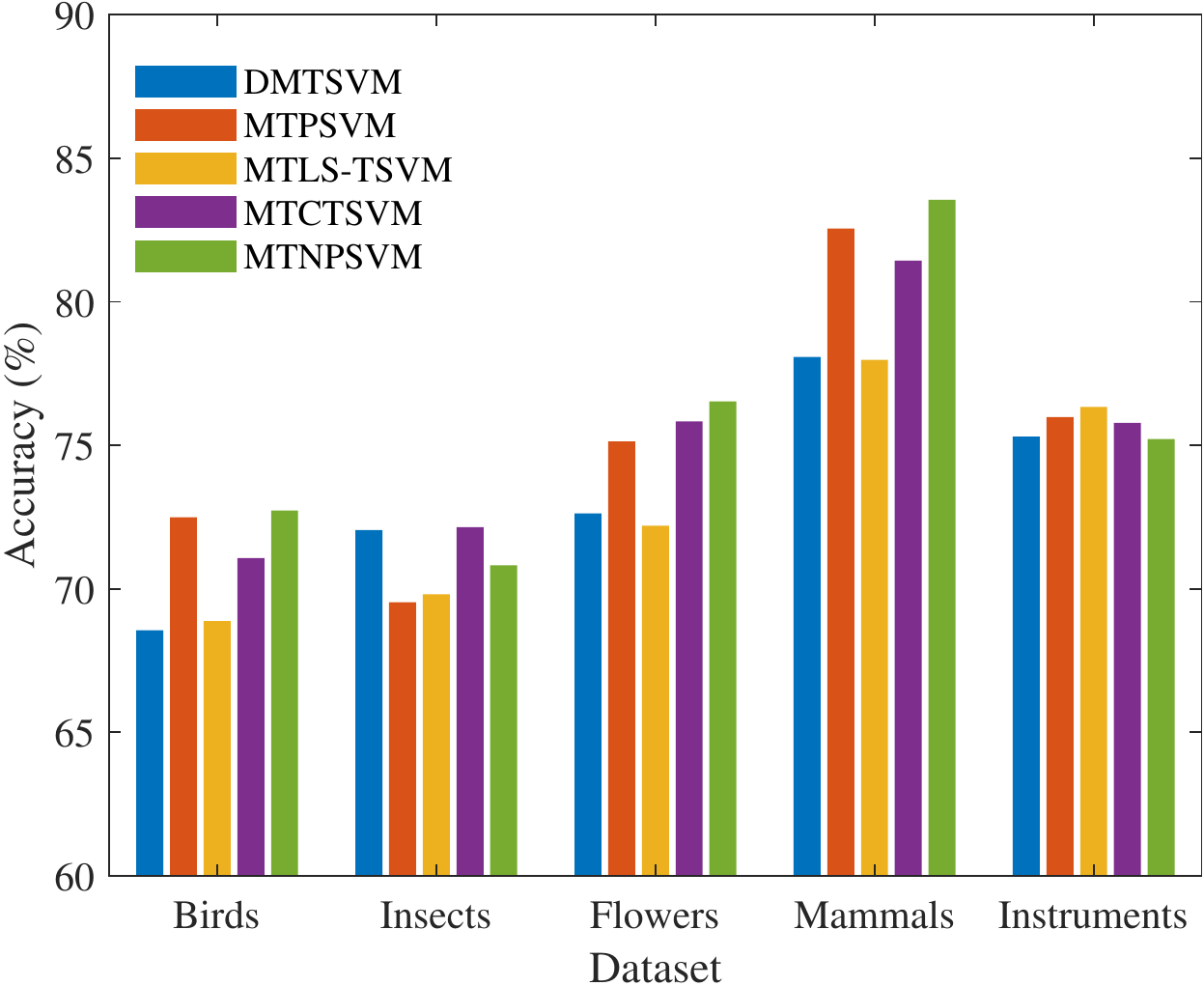}}
	\subfigure{\includegraphics[width=0.45\textwidth]{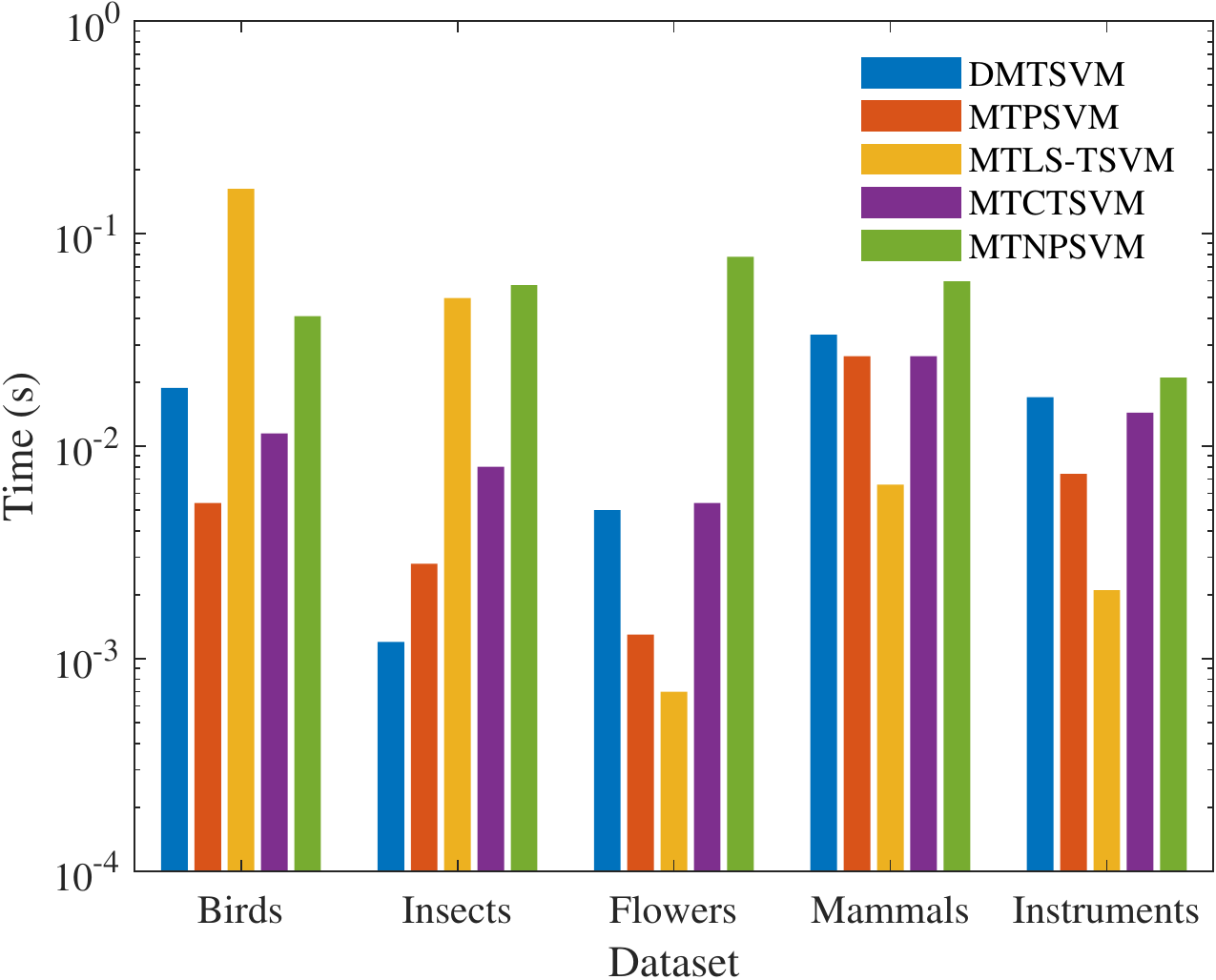}}	
	\caption{Performance comparison on Caltech 101 with RBF kernel.}	
	\label{cal101_guass}
\end{figure}

The dense-SIFT algorithm \cite{sift} is used for feature extraction. To further speed up the training efficiency based on retaining the original training information as much as possible, the PCA is introduced to reduce the original dimensions, while it can keep the original 97\% of the information. It should be noted here that the feature dimensions of the image datasets are still 300-600 dimensions by dimensionality reduction. Compared to the benchmark dataset, MTNPSVM does not perform very well in this case. In this subsection, the grid-search strategy and 5-fold cross-validation are also employed. The performance comparison on the five multi task datasets from Caltech 101 with RBF kernel are shown in Fig. \ref{cal101_guass}.

\begin{figure}[!htbp]	
	\centering
	\subfigure{\includegraphics[width=0.45\textwidth]{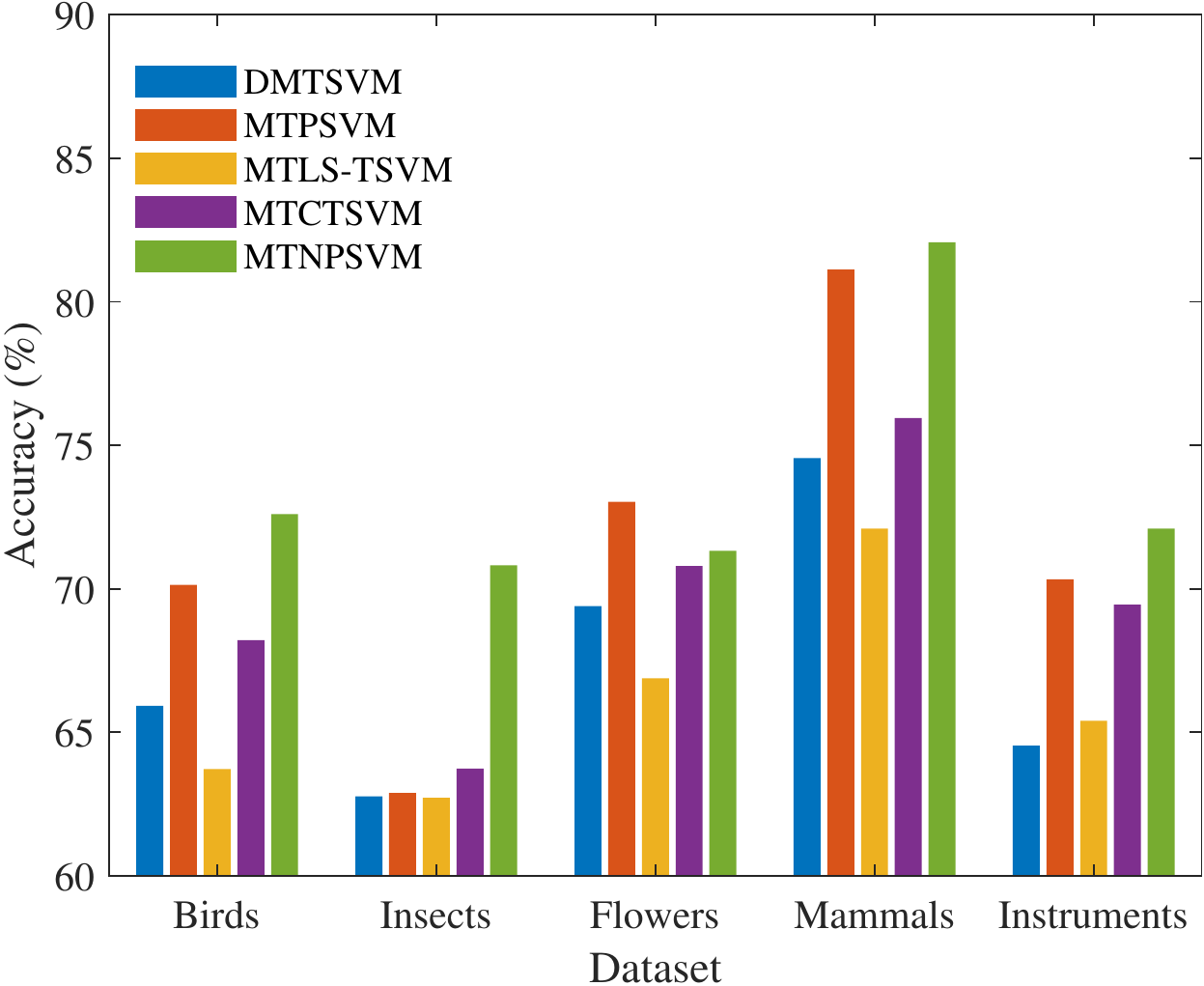}}
	\subfigure{\includegraphics[width=0.45\textwidth]{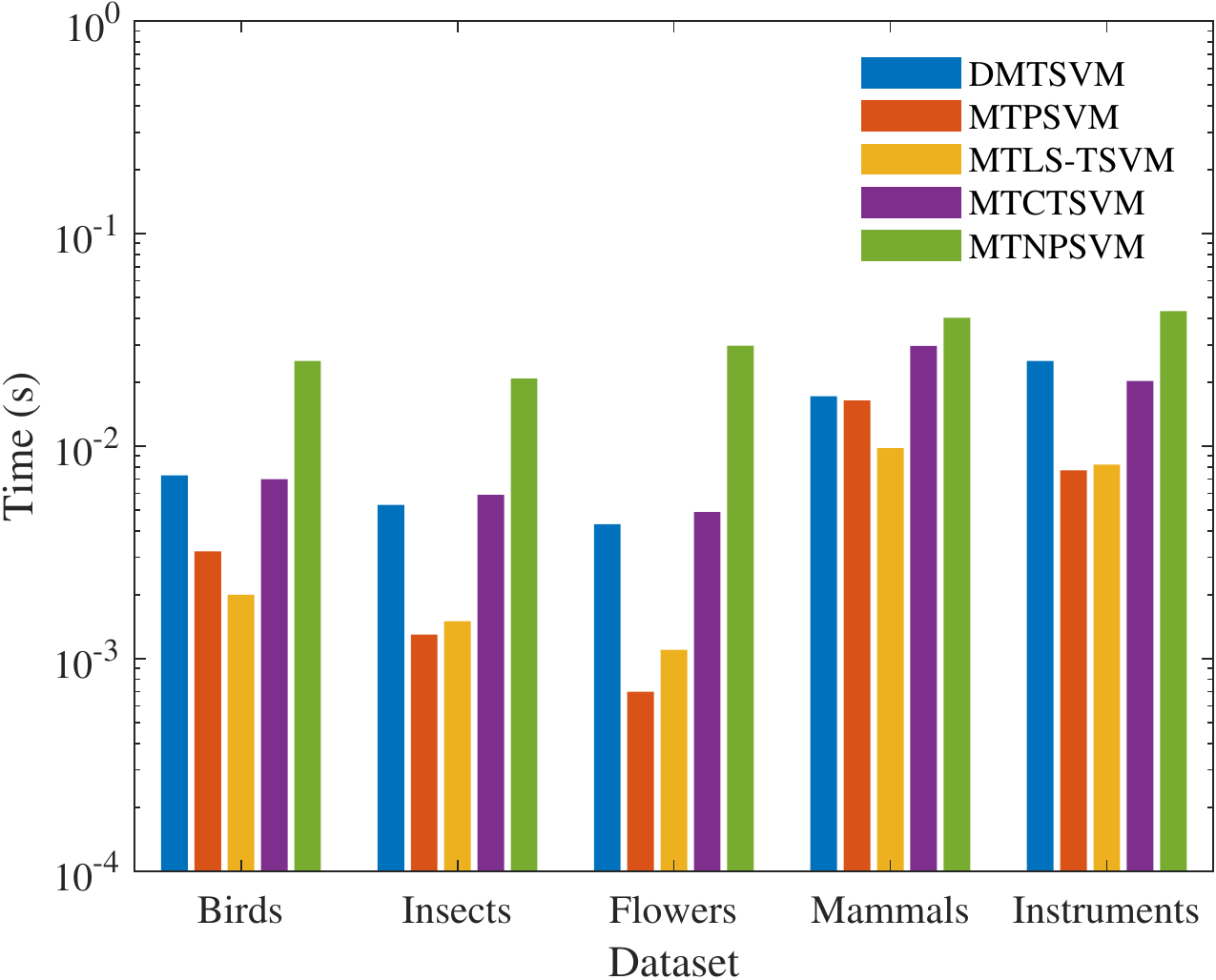}}	
	\caption{Performance comparison on Caltech 101 with Polynomial kernel.}	
	\label{cal101_poly}
\end{figure}

\begin{figure}[!htbp]	
	\centering
	\subfigure{\includegraphics[width=0.45\textwidth]{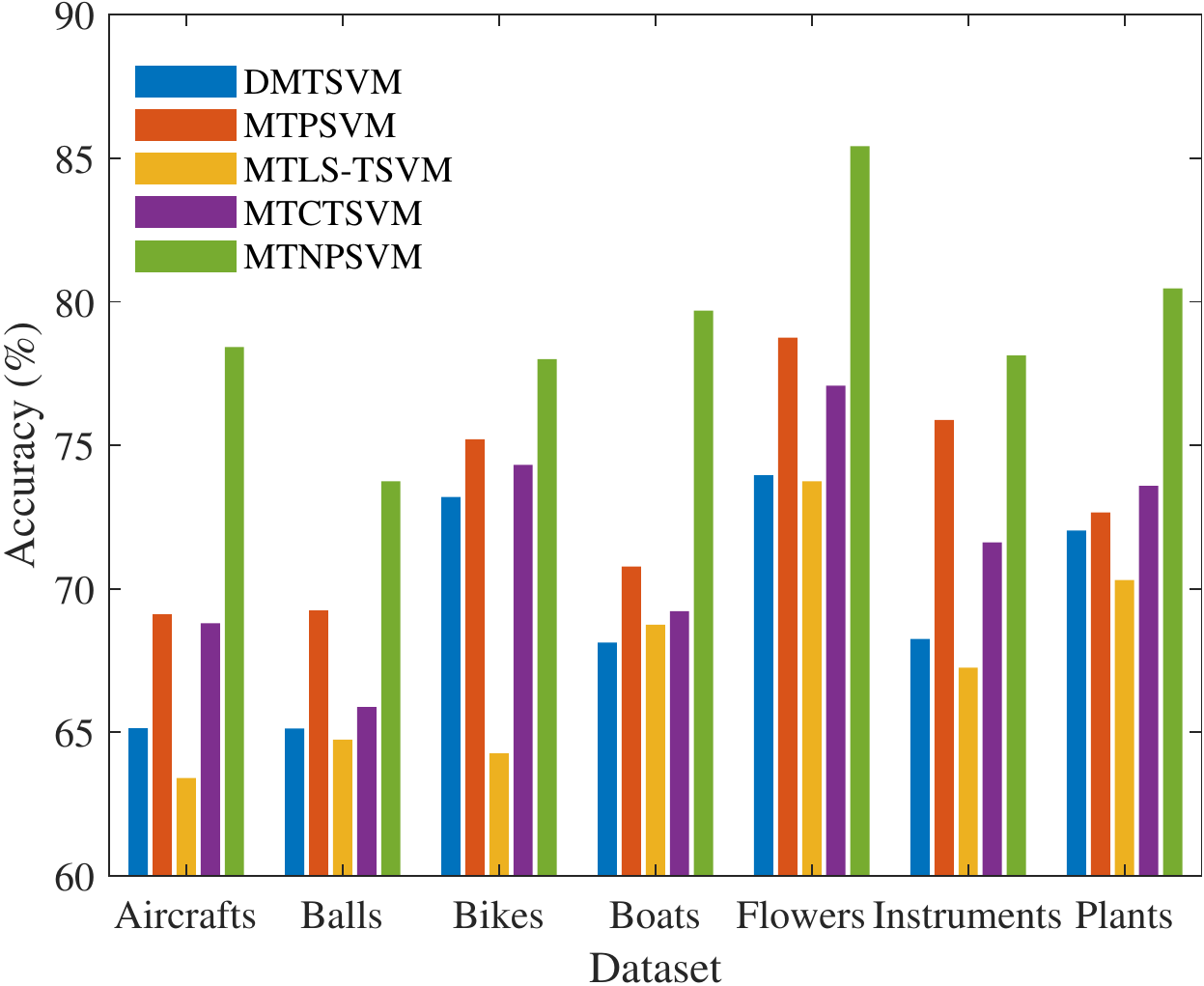}}
	\subfigure{\includegraphics[width=0.45\textwidth]{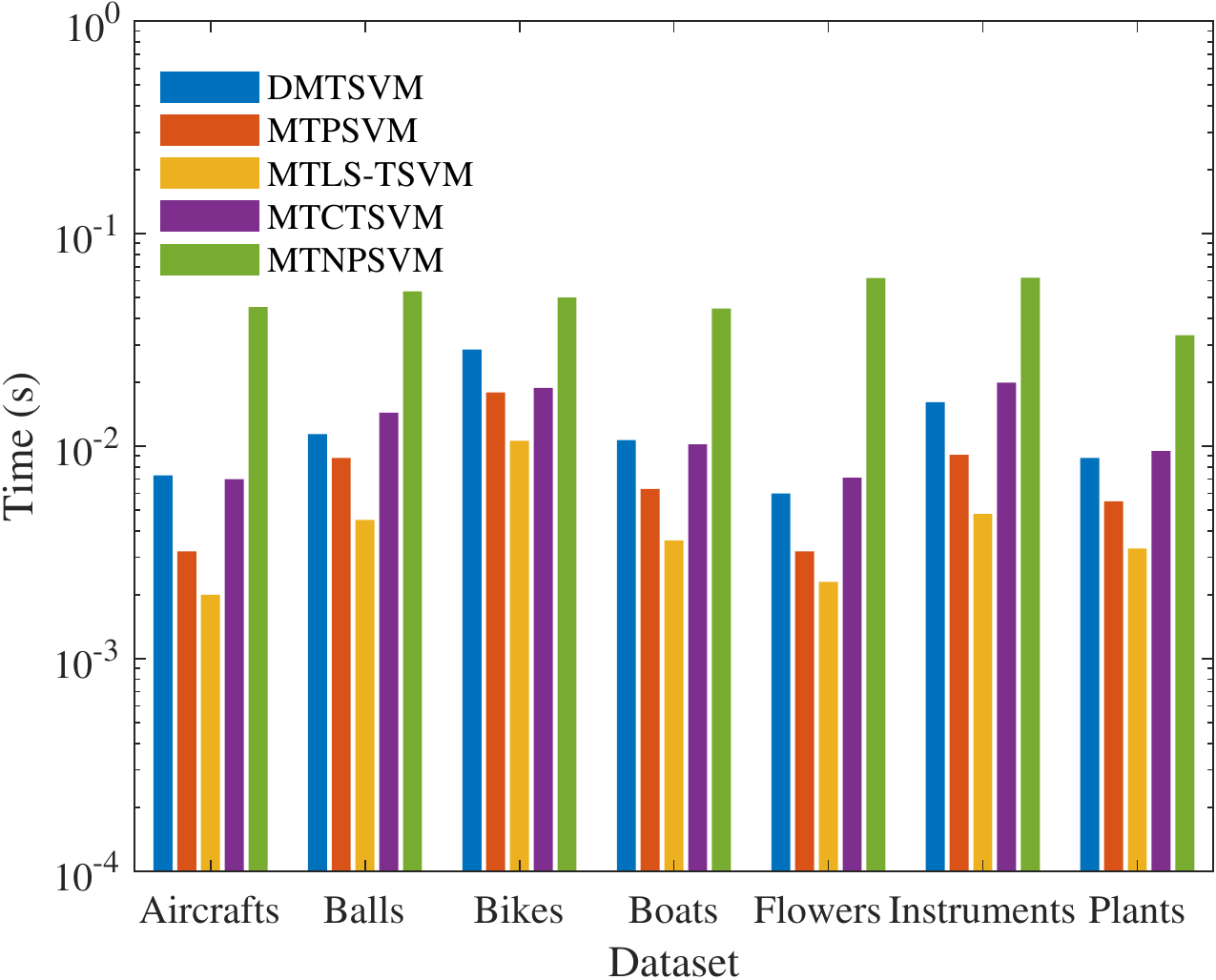}}	
	\caption{Performance comparison on Caltech 256 with Polynomial kernel.}	
	\label{cal256_poly}
\end{figure}

In terms of accuracy, the experimental results show that MTNPSVM performs slightly better than the other MTL methods. It can be explained as follows, the RBF kernel allows the samples to be mapped to a sufficiently high dimension, so that most of the samples can be linearly separable, thereby making the performance of all the models not easily distinguishable. In order to better reveal and compare the performance of the models, some experiments with Polynomial kernel are further implemented, which maps the features to a finite number of dimensions, the experimental results are displayed in Figs. \ref{cal101_poly} and \ref{cal256_poly}. Unlike the experiments results with RBF kernel, MTNPSVM can show more obvious advantages over other models, especially in seven datasets from Caltech 256. A similar statement of conclusion can also be drawn in the \cite{vtwin}.

In addition, in terms of computational time, since MTNPSVM requires the construction of a larger dimensional matrix, which results in more computational time. After acceleration by ADMM algorithm, the training time is still slightly higher than other models. Taking advantage of the high sparsity of the proposed model to improve the solving speed is the next research direction.
\subsection{Application in Chinese Wine}
From the numerical experiments above, it can be found that MTNPSVM has sufficient theoretical significance and good generalization performance because it inherits the common advantages of both NPSVM and multi-task learning. To further validate the practical significance of MTNPSVM, this subsection conducts comparable experiments with other models on the Chinese Wine dataset.

The wine dataset was collected from four areas, i.e., Hexi, Tonghua, Corridor, Helan Mountain, and Shacheng. Because the datasets from four different locations all have 1436 samples with 2203 feature dimensions, they can be considered as four highly related tasks. The grid-search strategy and 5-fold cross-validation are also performed in this dataset. By applying the above MTL methods with Polynomial kernel, the accuracies and optimal parameters used in experiment are displayed in Table \ref{table4}. After comparison, it can be found that MTNPSVM has better generalization performance than other multi-task models. In addition, it can be found that the parameter $\epsilon$ only has a large effect on the sparsity of the model, but has little effect on the prediction accuracy. Therefore it is suggested that the readers preset the parameter $\epsilon$ to 0.1 or 0.2. In this way, the added $\epsilon$ does not increase the burden of grid search.
\setlength{\tabcolsep}{2pt}
\begin{table*}[!htbp]	
	\centering
	\caption{The performance comparison on Chinese Wine dataset.}\label{table4}
	\begin{tabular}{cccccccccccc}
	 \hline
     Datasets&&\makecell{DMTSVM\\Accuracy(\%)\\(c,$\rho$,$\delta$)}&&\makecell[c]{MTLS-TSVM\\Accuracy(\%)\\(c,$\rho$,$\delta$)}&&\makecell[c]{MTPSVM\\Accuracy(\%)\\(c,$\rho$,$\delta$)}&&\makecell[c]{MTCTSVM\\Accuracy(\%)\\(c,$\rho$,$\delta$,$g$)}&&\makecell[c]{MTNPSVM\\Accuracy(\%)\\($c_{1}$,$c_{2}$,$\rho$,$\delta$,$\epsilon$)}\\
     \hline
      Chinese Wine&&\makecell[c]{74.81\\$(2^{-3},2,7)$}&&\makecell[c]{73.60\\$(2^{-3},2^{-3},1)$}&&\makecell[c]{77.42\\$(2^{-3},2^{2},4)$}&&\makecell[c]{73.60\\$(2^{-2},2^{-1},2^{-2},5)$}&&\makecell[c]{\textbf{78.62}\\$(2^{-3},2^{-3},2^{3},6,0.1)$}\\
     \hline
	\end{tabular}
\end{table*}
\section{Conclusion and further work}\label{10088}
This paper proposes a novel MTNPSVM which is an extension of the nonparallel support machine in the multi-task learning field. It both inherits the advantages of MTL and NPSVM, and overcomes the shortages of multi-task TWSVMs. It only needs to deal with one form of QPP for the linear case and the nonlinear case. Compared with the single task learning, MTNPSVM has a good generalization performance resulting from the task relations. Similarly, compared with the other MTL methods, MTNPSVM gets a better performance due to the introduction of the $\epsilon$-insensitive loss. Furthermore, it is proved that $\epsilon$ can flexibly adjust the sparsity of the model. Finally ADMM is introduced as the solving algorithm for the proposed model. Experiments on fifteen benchmark datasets and twelve image datasets are conducted to demonstrate the good performance of MTNPSVM. The application on the Chinese Wine dataset validates the practical significance of MTNPSVM. Combining the high full sparsity of the proposed model with algorithms to improve the solving rate is the future research direction.

\vspace*{10pt}
 {\bfseries
Acknowledgments} The authors gratefully acknowledge the helpful comments and suggestions of the reviewers, which have improved the
presentation. This work was supported in part by National Natural Science Foundation of China (No. 12071475, 11671010) and Beijing Natural Science Foundation (No. 4172035).

\appendix
\section{Proofs of Theorem \ref{th1}}\label{proof1}
 At the beginning, the KKT conditions for the primal problem are derived, in the main text a part of the KKT condition can be obtained by deriving the Lagrangian function as follows:
\begin{eqnarray}\label{a1}
	&&\rho_{1} u-\sum_{t=1}^{T} A_{t}^{\top}\left(\alpha_{t}^{+*}-\alpha_{t}^{+}\right)+\sum_{t=1}^{T} B_{t}^{\top} \beta_{t}^{-}=0, \\
	\label{a2}
	&&\frac{u_{t}}{T}-A_{t}^{\top}\left(\alpha_{t}^{+*}-\alpha_{t}^{+}\right)+B_{t}^{\top} \beta_{t}^{-}=0, \\
	&&C_{1} e_{1 t}-\alpha_{t}^{+}-\theta_{t}=0, \\
	&&C_{1} e_{1 t}-\alpha_{t}^{+*}-\psi_{t}=0, \\
	&&C_{2} e_{2 t}-\beta_{t}^{-}-\gamma_{t}=0.
\end{eqnarray}
In addition, one can get the following complementary relaxation conditions:
\begin{eqnarray}
	\label{a6}
   &&\alpha_{it}\left[\epsilon +\eta_{it}-\left(u+u_{t}\right)^{\top}x_{it}\right]=0,\\
   \label{a7}
   &&\alpha_{it}^{+*}\left[\epsilon +\eta_{it}^{*}+\left(u+u_{t}\right)^{\top}x_{it}\right]=0.
\end{eqnarray}
In order to prove Theorem \ref{th1}, the KKT conditions can be obtained by constructing the Lagrangian function of (\ref{dual1}) as follows:
\begin{eqnarray}
	&&M(A,A^{\top})(\alpha^{+*}-\alpha^{+})-M(A,B^{\top})\beta+\epsilon e^{+}-\overline{s}^{*}+\overline{\eta}^{*}=0,\\
	&&-M(A,A^{\top})(\alpha^{+*}-\alpha^{+})+M(A,B^{\top})\beta+\epsilon e^{+}-\overline{s}^{*}+\overline{\eta}^{*}=0,\\
	&&\overline{\eta}^{(*)}\ge0,\\
	&&\overline{s}^{(*)}\ge0,	
\end{eqnarray}
and
\begin{eqnarray}
	\label{a12}
	&&\overline{\eta}_{it}^{(*)}(C_{1}-\alpha_{it}^{+(*)})=0,\\
	&&\overline{s}_{it}^{(*)}\alpha_{it}^{+(*)}=0,
\end{eqnarray}
here the $\overline{\eta}^{(*)}$,$\overline{s}^{(*)}$ which are the new Lagrangian multipliers represent $\overline{\eta}$, $\overline{\eta}^{*}$ and $\overline{s}$, $\overline{s}^{*}$, respectively. The subscript letter $it$ of each vector represents the $i$ component of the $t$-th task. It should be mentioned that the $\overline{\eta}^{(*)}$ is equivalent to the relaxation variable ${\eta}^{(*)}$in the primal problem and also satisfies the equation (\ref{a6}) and (\ref{a7}). Detailed proof can be found in \cite{svr}. Now let us further discuss equations (\ref{a6}) and (\ref{a12}) to prove Theorem \ref{th1} in different situations.

if $\alpha_{it}^{+}\in(0,C_{1})$, According to (\ref{a6}) and (\ref{a12}), $\eta_{it}$=0, $(u+u_{t})x_{it}=\epsilon>-\epsilon$, further according to the constraints of the primal problem:
\begin{eqnarray}
	\label{a14}
	(u+u_{t})^{\top}x_{it}\ge-\epsilon-\eta_{it}^{+*},
\end{eqnarray}
the $\eta_{it}^{+*}$=0 can be obtained. By the (\ref{a7}), finally $\alpha_{it}^{+*}$=0 can be derived. Similarly, when $\alpha_{it}^{+*}\in(0,C_{1})$, it can also be prove that $\alpha_{it}^{+}$=0.

If $\alpha_{it}^{+}=C_{1}$, by the (\ref{a12}), $\eta_{it}^{+}\ge0$, from (\ref{a6}), the $(u+u_{t})x_{it}=\epsilon+\eta_{it}^{+}>-\epsilon$ can be obtained, further according to the (\ref{a14}), one can get $\eta_{it}^{+*}$=0, by the (\ref{a7}), finally $\alpha_{it}^{+*}$=0 can be derived. Similarly, when $\alpha_{it}^{+*}=C_{1}$, it can also be proved that $\alpha_{it}^{+}$=0.

Based on the above mentioned, it can be summarized that $\alpha_{it}^{+*}$$\alpha_{it}^{+}$=0. Theorem \ref{th1} is proved, and similarly the Theorem \ref{th3} can be proved by using problem (\ref{dual2}). They have the same proof procedure.
\section{Proofs of Theorem \ref{th2}}\label{proof2}
For the Theorem \ref{th2}, by following the KKT conditions (\ref{a1}) and (\ref{a2}), the equations can be converted into the following form:
\begin{eqnarray}
	&&u=\frac{1}{\rho_{1}}(\sum_{t=1}^{T} A_{t}^{\top}\left(\alpha_{t}^{+*}-\alpha_{t}^{+}\right)-\sum_{t=1}^{T} B_{t}^{\top} \beta_{t}^{-}),\\
	&&u_{t}=T(A_{t}^{\top}\left(\alpha_{t}^{+*}-\alpha_{t}^{+}\right)-B_{t}^{\top} \beta_{t}^{-}).
\end{eqnarray}
The same proof occurs in Theorem \ref{th4}.
\end{document}